\pdfoutput=1

\documentclass[11pt]{article}

\PassOptionsToPackage{table}{xcolor}
\usepackage[preprint]{acl}

\usepackage{times}
\usepackage{latexsym}

\usepackage[T1]{fontenc}

\usepackage[utf8]{inputenc}

\usepackage{microtype}

\usepackage{inconsolata}

\usepackage{hyperref}       
\usepackage{url}            
\usepackage{booktabs}       
\usepackage{amsfonts}       
\usepackage{nicefrac}       
\usepackage{microtype}      

\usepackage{booktabs}
\usepackage{ragged2e}
\usepackage{makecell}
\newcommand{\eat}[1]{}
\usepackage{subcaption}
\usepackage{amsmath}
\usepackage{enumitem}
\usepackage{tabularx}
\usepackage[pdftex]{graphicx}
\usepackage{color}

\newcommand{\todo}[1]{\textcolor{red}{[\textsc{TODO: }#1 ]}}

\newcommand{\prompt}[1]{{\tt \frenchspacing #1}}
\newcommand{\task}[1]{\textsc{\small #1}}
\newcommand{\evalstandardurl}{\url{https://github.com/allenai/olmes}}

\newcommand{\evalstandard}[0]{OLMES}
\newcommand{\evalstandardlong}[0]{Open Language Model Evaluation Standard}
\newcommand{\mc}[0]{MCF}
\newcommand{\rc}[0]{CF}

\usepackage{etoolbox}
\usepackage{pgf} 
\definecolor{highcolor}{HTML}{44ff44} 
\definecolor{midcolor}{HTML}{ffffff}  
\definecolor{lowcolor}{HTML}{ff0000}  
\newcommand*{\opacity}{70}
\newcommand*{\minval}{0.0}
\newcommand*{\midval}{50.0} 
\newcommand*{\maxval}{100.0}
\newcommand{\ccell}[1]{
    \ifdimcomp{#1pt}{>}{\maxval pt}{#1}{
        \ifdimcomp{#1pt}{<}{\minval pt}{#1}{
          \ifdimcomp{#1pt}{<}{\midval pt}{
            \pgfmathparse{int(round(100*(#1/(\midval-\minval))-(\minval*(100/(\midval-\minval)))))}\xdef\tempa{\pgfmathresult}\cellcolor{midcolor!\tempa!lowcolor!\opacity}#1}{
            \pgfmathparse{int(round(100*(#1/(\maxval-\midval))-(\midval*(100/(\maxval-\midval)))))}\xdef\tempa{\pgfmathresult}\cellcolor{highcolor!\tempa!midcolor!\opacity}#1}}}}

\definecolor{olmoBlue}{HTML}{265ed4}
\newcommand{\balpha}{{\color{olmoBlue}\boldsymbol{\alpha}}}
\newcommand{\bbeta}{{\color{olmoBlue}\boldsymbol{\beta}}}

\title{\evalstandard{}: A Standard for Language Model Evaluations}

\author{
Yuling Gu $^{\balpha}$
\quad  Oyvind Tafjord $^{\balpha}$
\quad  Bailey Kuehl $^{\balpha}$
\quad  Dany Haddad $^{\balpha}$ \\
{\bf Jesse Dodge $^{\balpha}$
\quad Hannaneh Hajishirzi $^{\balpha}$$^{\bbeta}$}\\
\\
$^{\balpha}$Allen Institute for Artificial Intelligence \quad
$^{\bbeta}$University of Washington \\
\texttt{\{yulingg, oyvindt\}@allenai.org}
}

\begin{document}
\maketitle
\begin{abstract}
Progress in AI is often demonstrated by new models claiming improved performance on tasks measuring model capabilities. Evaluating language models can be particularly challenging, as choices of how a model is evaluated on a task can lead to large changes in measured performance. There is no common standard setup, so different models are evaluated on the same tasks in different ways, leading to claims about which models perform best not being reproducible.
We propose \evalstandard{}, a completely documented, practical, open standard for reproducible LLM evaluations. 
In developing this standard, we identify and review the varying factors in evaluation practices adopted by the community – such as details of prompt formatting, choice of in-context examples, probability normalizations, and task formulation.
In particular, \evalstandard{} supports meaningful comparisons between smaller base models that require the unnatural ``cloze'' formulation of multiple-choice questions against larger models that can utilize the original formulation. 
\evalstandard{} includes well-considered, documented recommendations guided by results from existing literature as well as new experiments resolving open questions.\footnote{All prompts, examples, and code used for \evalstandard{} are available at \evalstandardurl{}.}
\end{abstract}

\section{Introduction}
\label{intro}
Scientific credibility in AI rests on reproducible and well-considered comparisons between models.
Many current AI models, such as pretrained large language models (LLMs), are generalist models capable of performing downstream tasks they were not specifically trained on \citep{GPT3NEURIPS2020, bommasani2022opportunities}. When evaluating LLMs on such tasks, there are many choices
in how the task is presented to the model and how the model outputs are interpreted before scoring \citep{eleuther-normalization, biderman2024lessons, liang2023holistic}. 
There is currently no standard way to decide on these choices, and they can have significant impact on model performance, with some recent papers claiming as much as an 80\% difference in accuracy on a given task just from varying formatting and in-context examples \citep{sclar2023quantifying}. 

\begin{table}[t]
\setlength\tabcolsep{1.5pt} 
  \centering
\begin{small}

\begin{tabular}{l|ccccccc}
\toprule
{\bf Model$\downarrow$} & {\bf Ref1} & {\bf Ref2} & {\bf Ref3} & {\bf Ref4} & {\bf Ref5} & {\bf Ref6} & {\bf OLMES} \\
\midrule
MPT-7B & \ccell{47.7} & \ccell{42.6} &  &  & \ccell{46.5} &  & \ccell{45.7}\\
RPJ-Incite-7B & \ccell{46.3} &  &  &  & \ccell{42.8} &  & \ccell{45.3}\\
Falcon-7B & \ccell{47.9} & \ccell{42.4} &  & \ccell{44.5} & \ccell{47.5} &  & \ccell{49.7}\\
Mistral-7B & \ccell{60.0} &  & \ccell{55.5} & \ccell{54.9} &  &  & $\ccell{78.6}^\dagger$ \\
Llama2-7B & \ccell{53.1} & \ccell{45.9} & \ccell{43.2} & \ccell{45.9} & \ccell{48.5} & $\ccell{53.7}^\dagger$ & \ccell{54.2} \\
Llama2-13B & \ccell{59.4} & \ccell{49.4} & \ccell{48.8} & \ccell{49.4} &  & $\ccell{67.6}^\dagger$ & $\ccell{67.3}^\dagger$\\
Llama3-8B & \ccell{60.2} &  &  &  &  & $\ccell{78.6}^\dagger$ & $\ccell{79.3}^\dagger$ \\
\midrule
Num shots & 25 & 0 & 0 & 0 & 0 & 25 & 5\\
Curated shots & No &  &  &  &  & No & Yes \\
Formulation & \rc{} & \rc{} & \rc{}? & \rc{} & \rc{} & \mc{} & \mc{}/\rc{} \\
Normalization & char & char & ? & char? & pmi & none & none/pmi\\
\bottomrule
\end{tabular}
\end{small}
\\
\vspace{2mm}
\begin{small}
\begin{tabular}{ll}
\toprule
\textbf{Ref} & \textbf{Reference citation} \\
\midrule
\textbf{Ref1} & HF Open LLM Leaderboard \citep{open-llm-leaderboard} \\
\textbf{Ref2} & Llama2 paper \citep{touvron2023llama}\\
\textbf{Ref3} & Mistral 7B \citep{jiang2023mistral}\\
\textbf{Ref4} & Falcon paper \citep{almazrouei2023falcon}\\
\textbf{Ref5} & OLMo paper \citep{groeneveld2024olmo}\\
\textbf{Ref6} & Llama3 model card \citep{llama3modelcard}\\
\bottomrule
\end{tabular} 
\caption{Scores reported in different references for LLM performances on \task{ARC-Challenge}. Scores indicated with $^\dagger$ are using multiple-choice formulation (\mc{}) rather than  ``cloze'' formulation (\rc{}) (see Section~\ref{sec:mcqa-llm} for definitions). Entries with ``?'' denote either undocumented or mixed approaches across models. Different references use different evaluation setups, some of which are not fully specified, so conclusions about performances and relative strengths of models are not reproducible.}
  \label{score-table-variations-arc-c}
\end{small}

\setlength\tabcolsep{6pt} 
\end{table}

These choices in evaluation setups are often not reported with enough details to reproduce, so when a team of ML practitioners releases a new model it is often impossible for them to directly compare against previously-reported results by others. Efforts like the Holistic Evaluation of Language Models (HELM) benchmark \citep{liang2023holistic} and 
the Hugging Face Open LLM Leaderboard \citep{open-llm-leaderboard} tackle the issue of reproducibility by striving towards standardizing LM evaluations. While the same setup is used to evaluate many models ensuring consistency and reproducibility, the rationale behind prompt formatting, use of in-content examples, normalization techniques, and task formulation are not always clearly documented and thus not consistently followed by other researchers in subsequent work \citep{touvron2023llama2,biderman2023pythia,jiang2023mistral,groeneveld2024olmo,llama3modelcard}. 

We highlight two problems in the field today.
(1) Releasing a new model and comparing it against previously reported results is flawed unless the previous work explicitly described their full evaluation setup, and then that setup is followed in the new work. Currently, different references (like leaderboards, papers) use different (and sometimes under-specified) evaluation setups, leading to different results and conclusions. For a particular dataset, evaluating a specified language model, different references can tell you very different stories.
We illustrate this phenomenon in Table~\ref{score-table-variations-arc-c}, which shows how several models' published performance on the \task{ARC-Challenge} \citep{clark2018think} task can vary in the literature. For instance, looking at Ref1, we would conclude that Llama2-13B and Llama3-8B are performing similarly, but Ref6 reveals there is likely a gap of over 10\% between them. 
(2) Despite current efforts to standardize model evaluation (e.g., HF Open LLM Leaderboard, HELM), the choices made are not justified and most model creators do not use these setups for their evaluations. We also see evidence of this in Table~\ref{score-table-variations-arc-c}, showing a variety of setups being used for the same task, differing in choices such as number of shots, source of in-context examples, task formulation, and probability normalization. While these different choices are made in implementing the evaluations, to date, there is no documented standard studying and/or justifying if one choice is better than another, leading to the lack of a set of justified choices that the community can adopt.  \citet{helm-mmlu} also demonstrates how this might be a community-wide problem in a recent effort (see Figure~\ref{fig:helm_mmlu_reproduction} in Appendix).

To address these problems, we present \evalstandard{} (\evalstandardlong{}), a standard to improve the transparency and reproducibility of language model evaluation from a practical point of view, removing ambiguity in how a final performance metric is obtained when evaluating a model on a dataset. \evalstandard{} can be applied to evaluation during the model development process, and in published leaderboards and papers. \evalstandard{} provides justified recommendations on all aspects of task setups, such as data sampling, how to format instances, the choice of in-context examples, probability normalization, and task formulation.

Importantly, \evalstandard{} is:
\begin{itemize}
\itemsep0em 
\item {\bf Reproducible:} \evalstandard{} specifies all details of the evaluations, from processing datasets to presenting the task to model, to processing models' outputs, so there are no ambiguities in the evaluation procedure.
\item {\bf Practical:} \evalstandard{} makes practical decisions in use of computation resources for easy adoption by the community.
\item {\bf Documented:} Each decision in the standard is documented with justifications by applying principles from existing studies and performing experiments to resolve open questions.
\item {\bf Open}: We release all prompts and code, along with the rationales behind the choices made in \evalstandard{}, for subsequent work to follow and build upon by extending the same principles to any new task and model.
\end{itemize}

Since \evalstandard{} is a documented, practical, open evaluation standard, it is straightforward to adopt in publicly available, well-maintained evaluation code bases like the Eleuther LM Evaluation Harness \citep{eval-harness, biderman2024lessons} and HELM \citep{liang2023holistic}. When used by model developers and other researchers, \evalstandard{} will help unify evaluation practices in the field. We believe this work is the first of its kind to unify practices for evaluating base models throughout the full development cycle, from small to large models as well as early to late training stages. All prompts, examples, and code used for \evalstandard{} can be found at \evalstandardurl{}.

\section{Experimental setup}
\label{experimental-setup}
\subsection{Multiple-choice QA and LLM evaluation}
\label{sec:mcqa-llm}
Multiple-choice question answering (MCQA) tasks present a  compelling way of evaluating models and humans alike,
due to the ease of scoring (whether the correct answer is chosen out of the given options) and the allowed flexibility in the domain and complexity of the questions. One motivation for multiple-choice tasks is that early in training, and for smaller base models before instruction-tuning, other tasks (generative tasks, math reasoning, coding, etc) tend to provide less useful signals. Multiple-choice tasks are the most common type of benchmarks for evaluating base LLMs \citep{open-llm-leaderboard, touvron2023llama, jiang2023mistral, groeneveld2024olmo, llama3modelcard}, where the evaluation seems straightforward (did the model predict the right answer?), but in practice, a statement like ``model X scores Y on \task{ARC-Challenge}'' is generally uninterpretable (with unspecified details and cannot be meaningfully compared across references, see Table~\ref{score-table-variations-arc-c}) without a clear evaluation standard like \evalstandard{}.

We specifically focus on evaluation using these tasks to provide useful guidance during and after base model training, giving important insights into the potential of such models before committing to further tuning (e.g., instruction-tuning). Such tasks form a large, essential part of LLM evaluations and are the focus of \evalstandard{}. There are generally two ways to formulate these tasks.

\paragraph{\bf \mc{} (Multiple-choice formulation):} presenting answer choices indicated by labels and scoring prediction of answer labels, just like how MCQA is posed to humans. Here is an example of MCQA from \task{ARC-Easy} \citep{clark2018think}, a dataset of 
real grade-school level science questions: 

\noindent{\tt \small \frenchspacing Question: Earth's core is primarily composed of which of the following materials?\\
\hspace*{0.7em}A. basalt\\
\hspace*{0.7em}B. iron\\
\hspace*{0.7em}C. magma\\
\hspace*{0.7em}D. quartz \\
Answer: B}
\paragraph 
{\bf \rc{} (Completion/cloze formulation):} scoring each answer choice separately using LLM token probabilities. The \mc{} format is not natural for the pure language modeling task of generating the next token. Therefore, the \rc{} format was introduced when evaluating the GPT-3 model \citep{GPT3NEURIPS2020}. They found that it was possible to elicit much better performance using a ``cloze'' completion version of the task, where the model is shown a prompt like:

\noindent{\tt \small \frenchspacing Question: Earth's core is primarily composed of which of the following materials? \\
Answer: <answer>}

Each answer choice is separately substituted in for {\tt \small <answer>}. Then the LLM probability of the answer choice tokens are used to rank the choices and predict an answer. This formulation has ambiguities in how to normalize the probability, as well as absolute limitations, such as not being able to properly address cases where one answer choice is ``none of the above'' or similar.

\subsection{Targeted tasks}

\begin{table*}[h]
  \centering

\begin{small}
\begin{tabular}{llllll}
\toprule
\bf{task} & \bf{split} & \bf{\#C} & \bf{\# inst (total)} & \bf{\rc\ norm} & \bf{reference}\\
\midrule
\task{ARC-Challenge} (ARC\_C) & Test & $4^\dagger$ & 1172 & pmi & \citep{clark2018think} \\
\task{ARC-Easy} (ARC\_E) & Test & $4^\dagger$ & 1000 (2376) & char & \citep{clark2018think}\\
\task{BoolQ} & Val & 2 & 1000 (3270) & none & \citep{clark-etal-2019-boolq}\\
\task{CommonsenseQA} (CSQA) & Val & 5 & 1221 & pmi & \citep{talmor-etal-2019-commonsenseqa} \\
\task{HellaSwag} (HSwag) & Val & 4 & 1000 (10042) & char & \citep{zellers-etal-2019-hellaswag}\\
\task{MMLU} & Test & 4 & 14042 & char & \citep{hendryckstest2021}\\
\task{OpenbookQA} (OBQA) & Test & 4 & 500 & pmi & \citep{mihaylov-etal-2018-suit} \\
\task{PIQA} & Val & 2 & 1000 (1838) & char & \citep{Bisk_Zellers_Le_bras_Gao_Choi_2020}\\
\task{Social IQa} (SIQA) & Val & 3 & 1000 (1954) & char & \citep{sap-etal-2019-social}\\
\task{WinoGrande} (WinoG) & Val & 2 & 1267 & none & \citep{Sakaguchi_Le_Bras_Bhagavatula_Choi_2020}\\
\bottomrule
\end{tabular}
\end{small}
  \caption{{\bf \evalstandard{}} details on tasks, with our standardized choices of dataset split, number of instances to use (along with total number if sampling was used), and which \rc{} normalization scheme to use (see Section~\ref{different_normalization}). Column {\bf\#C} shows the number of answer choices (\task{ARC-Challenge} and \task{ARC-Easy}$^\dagger$ have a few instances with 3 or 5 answer choices). See Section~\ref{sec:standardizing-variatons-main} for details on instance formatting, choice of in-context examples and task formulation.}
  \label{task-details}
\end{table*}

We select and implement standards for 10 popular benchmark MCQA tasks, see Table~\ref{task-details} for the list. The list covers tasks that are frequently used in the community's evaluation practices, such as the Hugging Face Open LLM Leaderboard \citep{open-llm-leaderboard}, Llama papers \citep{touvron2023llama, touvron2023llama2, llama3modelcard},  HELM \citep{liang2023holistic}, and the OLMo evaluation suite \citep{groeneveld2024olmo}. This selection includes questions on science, various types of commonsense, factual knowledge, and covers a range of topics (\task{MMLU} alone covers 57 subjects), of varying difficulty.

\subsection{Selection of models}
We develop \evalstandard{} based on a selection of 15 diverse, openly available pretrained LLMs, focusing on base (not instruction-tuned) models, covering a range of sizes from 1B to 70B -- Pythia-1B, Pythia-6.7B \citep{biderman2023pythia}, OLMo-1B, OLMo-7B, OLMo-7B-0424 \citep{groeneveld2024olmo}, TinyLlama-1.1B \citep{zhang2024tinyllama}, StableLM2-1.6B \citep{bellagente2024stable}, RPJ-INCITE-7B \citep{together2023redpajama},  MPT-7b \citep{mosaic-mpt}, Falcon-7B \citep{almazrouei2023falcon}, Llama2-7B, Llama2-13B \citep{touvron2023llama2}, Mistral-7B-v0.1 \citep{jiang2023mistral}, Llama3-8B, Llama3-70B \citep{llama3modelcard}. 
This reflects our goal of providing an evaluation standard that suits a range of model capabilities, with the flexibility to apply the same methodology during model development as well as when comparing final powerful base models.

Assessing base models of different strengths is important during the training of models and before it is used for further tuning (e.g., instruction-tuning). This is critical for the community when picking between alternate base models for further training or tuning for their application. There is limited established protocol in the community – evaluation during training is often left underspecified and understudied, and when evaluating final base models, researchers across the field use different evaluation setups, leading to different results and conclusions (Tables~\ref{score-table-variations-arc-c} and \ref{score-table-variations-obqa}). We hope this work will empower the community towards more unified practices in benchmarking base models so that further progress can be made on a stronger foundation based on careful evaluation.

\section{Standardizing variations in evaluation}
\label{sec:standardizing-variatons-main}
To evaluate a model on a dataset, there are a variety of decisions that have to be made to get a final score of that model on that dataset. These include:

\begin{itemize}
\itemsep0em
\item How to format dataset instances? (Section~\ref{sec:varied-dataset-use})
\item Which few-shot examples to use? (Section~\ref{section-varied-shots})
\item How to normalize LLM probabilities for \rc{}? (Section~\ref{different_normalization})
\item What task formulation to use, MCF or CF? (Section~\ref{sec:varied-format-mc-vs-rc}) 
\item Other implementation choices impacting results (Section~\ref{sec:other-specifications})
\end{itemize}

Below we enumerate key variations in these steps, and justify the choices made in \evalstandard{} (some of which are summarized in Table~\ref{task-details}) to standardize these steps, leaving some of the details for the Appendix.

\subsection{How to format dataset instances?}  
\label{sec:varied-dataset-use}
Each MCQA dataset includes a set of fields used to specify an instance, such as question, answer choices, and perhaps a context for the question. When formatting an instance as a prompt to an LLM, many different choices have been made in the literature. This includes simple choices like \prompt{"Question:"} vs \prompt{"Q:"} as question prefix (varying even within a paper, e.g., \citet{GPT3NEURIPS2020}), or formatting the answer labels (e.g., \prompt{"A."} \citep{touvron2023llama}, \prompt{"(A)"} \citep{nori2023capabilities}, \prompt{"<mc>A</mc>"} \citep{claude}, etc). There is also a choice of whether or not to provide a general instruction, e.g., common for \task{MMLU} \citep{hendryckstest2021}, sometimes done for \task{OpenbookQA} \citep{almazrouei2023falcon}.

\noindent\textbf{Instance formatting.} \textbf{\evalstandard{}} uses a consistent \prompt{"Question: <question>"} prefix and {\tt \frenchspacing "Answer:"} suffix in formatting the datasets. This clarifies the question-answering task in a natural way, without relying on verbose instruction understanding. The three exceptions are listed and explained here. For \task{PIQA}, we use \prompt{"Goal: <goal>"} as the prefix instead to be consistent with the original semantics of the dataset. In the case of \mc{}, for \task{HellaSwag}, we skip the question prefix and instead add \prompt{"Choose the best continuation:"} before presenting the continuation options, and for \task{WinoGrande} we use the prefix \prompt{"Fill in the blank:"} to align with the task. For \task{HellaSwag} and \task{WinoGrande}, where the \rc{} answer string is simply a language continuation, we remove such prefixes and suffixes for the \rc{} evaluation so that the task is closer to pure language modeling.

\noindent\textbf{MCQA label choice.} For \mc{} answer choices, \evalstandard{} uses the canonical letters A/B/C/… as answer labels, presenting the multiple-choice options after simple letter labels, i.e., \prompt{" A."} format. We note that most tokenizers treat a letter at the start of a line (or string) as a separate token from the same letter following a space. Therefore we add a prefix space in front of each answer label \prompt{"\textbackslash n~A.~<choice>"} (rather than \prompt{"\textbackslash nA.~<choice>"}), to work naturally with all current tokenizers (so that the final answer token will be identical to the answer choice token, see Appendix~\ref{tokenization-details-appendix} for details). All the exact \evalstandard{} prompt formats are listed in Appendix~\ref{task-prompt-formats-appendix}.

\noindent\textbf{Sampling.} Following existing LLM evaluation standardization efforts \citep{liang2023holistic, open-llm-leaderboard}, \textbf{\evalstandard{}} uses the test split of a dataset if the labels are publicly available, otherwise the validation split. If the dataset has more than 1500 instances, we sample 1000 instances to evaluate,\footnote{Sampling uses a specific random seed in Python: \prompt{Random(1234).sample(all\_instances, 1000)}} similar to HELM \citep{liang2023holistic} which caps evaluation instances at 1000.\footnote{\url{https://crfm-helm.readthedocs.io/en/latest/reproducing_leaderboards/}} Note that the potential extra statistical signal from more instances would generally be dominated by other sources of score variations, like prompt formatting, so this is a practical consideration to avoid unnecessary computation resources. See Table~\ref{task-details} for details on splits and sampling used in \evalstandard{}.

\subsection{Which few-shot examples to use?}
\label{section-varied-shots}

Popularized by \citet{GPT3NEURIPS2020}, it is customary to provide examples of the task to the model through few-shot examples, as this is an effective and universal way to convey a task to an LLM. For example, the \task{MMLU} task \citep{hendryckstest2021} originally came with a fixed 5-shot prompt which is generally used in evaluation \citep{open-llm-leaderboard, gemmateam2024gemma, jiang2023mistral, touvron2023llama2, llama3modelcard} resulting in more reproducible results than many other tasks.\footnote{Sometimes sampled examples are used also for \task{MMLU} \citep{mosaic-eval-suite}. Even for \task{MMLU}, noticeable discrepancies have been found, due to other differences in prompt formatting \citep{helm-mmlu}.}
For other tasks, both the number of shots and the way in which they are sampled have varied in different evaluation setups. For example, to evaluate on \task{HellaSwag}, \citet{open-llm-leaderboard} sampled 10-shot whereas HELM \citep{liang2023holistic} uses 0-shot; within \citet{open-llm-leaderboard}, a range of 25-shot, 10-shot, 5-shot was sampled for \task{ARC-Challenge}, \task{HellaSwag} and \task{WinoGrande} respectively.

\textbf{\evalstandard{}} standardizes a manually curated 5-shots prompt for each task (from its training set), ensuring that the examples are of good quality and cover the label space in a balanced way (e.g., avoiding 4 A's and 1 B among the 5 answers).\footnote{More details on curating the examples can be found in Appendix~\ref{curation_5_shot_considerations}.} Restricting to 5 in-context examples helps limit computational overhead, similar to HELM \citep{liang2023holistic}. Analysis suggests that going beyond 5 shots generally does not provide meaningful differences in scores \citep{GPT3NEURIPS2020, mosaic-eval-gauntlet}. The manually curated shots for each task can be downloaded from \evalstandardurl{}.

\subsection{How to normalize LLM probabilities for \rc{}?}
\label{different_normalization}

When using the completion/cloze formulation (\rc{}) for multiple-choice questions, the LLM returns $P(a_i|q)$, the probability for an answer choice $a_i$ given a question prompt $q$. Ranking solely based on the probability may heavily favor shorter answers with fewer tokens. To work around this issue, different normalization methods have been used in the literature, which we categorize below: 

\begin{itemize}[leftmargin=5mm]
\itemsep0em
\item \textbf{none:} $\ln(P(a_i|q))$
\item \textbf{token:} $\ln(P(a_i|q)) / \operatorname{num\_tokens}(a_i)$, which normalizes the log-probability by the number of tokens in the answer~\citep{GPT3NEURIPS2020}.
\item \textbf{character:} $\ln(P(a_i|q)) / \operatorname{num\_characters}(a_i)$, which normalizes the log-probability by the number of characters in the answer, used by Llama models \citep{touvron2023llama} and Eleuther AI LM Harness~\citep{eval-harness, biderman2024lessons}.
\item \textbf{pmi:} $\ln(P(a_i|q) / P(a_i|u))$ where $u=${\tt \frenchspacing "Answer:"} is an unconditional prompt, which normalizes by dividing by the LLM probability of the same answer string without the presence of the question. This can be considered a form of pointwise-mutual-information (PMI) and was explored further in other works \citep{holtzman-etal-2021-surface}.
\end{itemize}

Efforts like \citet{liang2023holistic, eval-harness, biderman2024lessons} compare and support comparisons of different normalization approaches, leaving it an open question as to how to make a decision. See Appendix~\ref{normalization-details-appendix} for further discussions around different normalizations.

To choose a normalization scheme in \textbf{\evalstandard{}}, we evaluate the models on each dataset, comparing the effect of the 4 normalization techniques. Table~\ref{tab:rc-norm-comparison-summary} shows for each task, how often each normalization is empirically the best across the 15 models. Detailed scores per model are in Appendix~\ref{normalization-details-appendix}.

\begin{table}[t!]
\setlength\tabcolsep{1.5pt}
  \centering
\begin{small}
\begin{tabular}{lcccc|cc}
\toprule
 & \multicolumn{4}{c|}{\bf win percentage} & diff &\\
task & none & char & tok & pmi & oracle & {\bf OLMES}\\
\midrule
ARC\_C & 0.0 & 33.3 & 0.0 & 66.7 & 0.2 & pmi\\
ARC\_E & 6.7 & 86.7 & 6.7 & 0.0 & 0.1 & char\\
BoolQ & 46.7 & 46.7 & 0.0 & 6.7 & 1.1 & none\\
CSQA & 6.7 & 33.3 & 6.7 & 53.3 & 0.6 & pmi\\
HSwag & 0.0 & 100.0 & 0.0 & 0.0 & 0.0 & char\\
MMLU & 0.0 & 46.7 & 0.0 & 53.3 & 0.4 & char\\
OBQA & 0.0 & 0.0 & 0.0 & 100.0 & 0.0 & pmi\\
PIQA & 6.7 & 46.7 & 46.7 & 0.0 & 0.2 & char\\
SIQA & 0.0 & 86.7 & 6.7 & 6.7 & 0.1 & char\\
WinoG & 100.0 & 0.0 & 0.0 & 0.0 & 0.0 & none\\
\bottomrule
\end{tabular}
\end{small}
\setlength\tabcolsep{6pt} 
  \captionof{table}{Summary of \rc{} normalization comparisons. ``win percentage'' shows how often each normalization was best across the 15 models. ``diff oracle'' (difference between the OLMES recommendation and the empirically best normalization for each task and model) shows that there is in general minimal difference between the \evalstandard{} normalization and the oracle optimal normalization for each task (difference out of 100\%).}%
  \label{tab:rc-norm-comparison-summary}
\end{table}

\evalstandard{} specifies the \textbf{``pmi''} normalization for \textsc{\small ARC-Challenge}, \task{CommonsenseQA}, and \task{OpenbookQA}. The answer choices in these datasets tend to contain unexpected words or phrases that are less likely for models to generate  (e.g., ``Whirlpool bath'' compared to ``Bathtub''). The pmi normalization adjusts for this by taking into account the a priori likelihood of the answers. This is consistent with other findings \citep{holtzman-etal-2021-surface} and some existing evaluation practices, e.g., \citet{GPT3NEURIPS2020} selectively uses this normalization for \task{ARC} and \task{OpenbookQA}, and \citet{touvron2023llama,touvron2023llama2} for \task{OpenbookQA}. Computing the extra unconditional likelihood incurs some computation overhead, thus \evalstandard{} avoids this normalization for other datasets where there is no strong empirical or theoretical reason to choose this approach.

\evalstandard{} specifies the \textbf{``character''} normalization for \task{ARC-Easy}, \task{HellaSwag}, \task{PIQA}, \task{Social IQa} and \task{MMLU}. 
Based on our experiments, it is empirically the normalization technique that gives the best scores\footnote{Tie for \task{PIQA}, and second-best for \task{MMLU}.}  for these datasets, and less computationally expensive than the ``pmi'' normalization. It also has the advantage (unlike the ``token'' normalization) of already being implemented (as {\tt acc\_norm}) in the Eleuther LM Evaluation Harness, where it is generally available for multiple-choice tasks \citep{eval-harness,eleuther-normalization}. It is also used in the Hugging Face Open LLM Leaderboard \citep{open-llm-leaderboard} for \task{ARC-Challenge} and \task{HellaSwag}, in \citet{touvron2023llama,touvron2023llama2}'s evaluations as the default, (with select datasets as exceptions), as well as reported in various works like \citet{biderman2023pythia,almazrouei2023falcon}. 

\evalstandard{} specifies the \textbf{``none''} normalization for \task{BoolQ} and \task{WinoGrande}. In \task{BoolQ} the only answer choices are ``yes'' or ``no'' which are single tokens, therefore no length normalization is needed. Note that for some models, the ``character'' normalization has slightly better performance on \task{BoolQ} (see Table \ref{tab:appendix-rc-norm-1}), an accidental side effect of ``yes'' having one more character than ``no''. One could argue that the pmi normalization is appropriate as it counters any existing bias in the model for ``yes'' vs ``no'', but we argue that models should be capable of producing such common words (also indicated in the 5-shot examples) without any such corrections. Finally, \task{WinoGrande} is a special case in that the continuations are identical (and the prompts vary), so the choice of normalization does not matter and we simply use the \textbf{``none''} normalization. 

In general, we observe little difference between the \evalstandard{} recommendation and the empirically best (``oracle'') normalization for each task and model, see ``diff oracle'' column in Table~\ref{tab:rc-norm-comparison-summary} (Table~\ref{tab-rc-norm-diff} in Appendix~\ref{normalization-details-appendix} has more details).

\subsection{What task formulation to use, \mc{} or \rc{}?}
\label{sec:varied-format-mc-vs-rc}

As LLMs have gotten stronger, the MCQA task formats have gradually changed from \rc{} to \mc{}. For instance, \task{ARC-Challenge} was often evaluated using the \rc{} approach \citep{touvron2023llama,touvron2023llama2,almazrouei2023falcon,open-llm-leaderboard}, but has switched to \mc{} for stronger models like \citet{openai2024gpt4,llama3modelcard}, appearing with identical names like ``25-shot \task{ARC-Challenge}''. As an example, \citet{llama3modelcard} reports a 25-shot \mc{} \task{ARC-Challenge} score for the Llama-3 8B model of 78.6\% vs 60.2\% for the 25-shot \rc{} on the Hugging Face Open LLM Leaderboard. As performance on a multiple-choice task gets closer to 100\%, the \rc{} approach lags behind due to its inherent limitations, giving significantly less signal about a model's actual performance. On the other hand,  \task{MMLU} is almost exclusively evaluated using the \mc{} approach, which often results in near-random performance for weaker models \citep{open-llm-leaderboard}. 

In \textbf{\evalstandard{}}, we argue that the \rc{} formulation provides a useful evaluation of task knowledge for models that have not yet acquired the skill of answering multiple-choice questions using \mc{}. 
On the other hand, \mc{} is a more realistic formulation for models that can ``understand'' this format, yielding higher and more representative scores \citep{robinson2023leveraging, openai2024gpt4}. See Appendix~\ref{task-formulation-details-appendix} for further discussion.

\begin{figure}[t!]
      \includegraphics[width=\columnwidth]{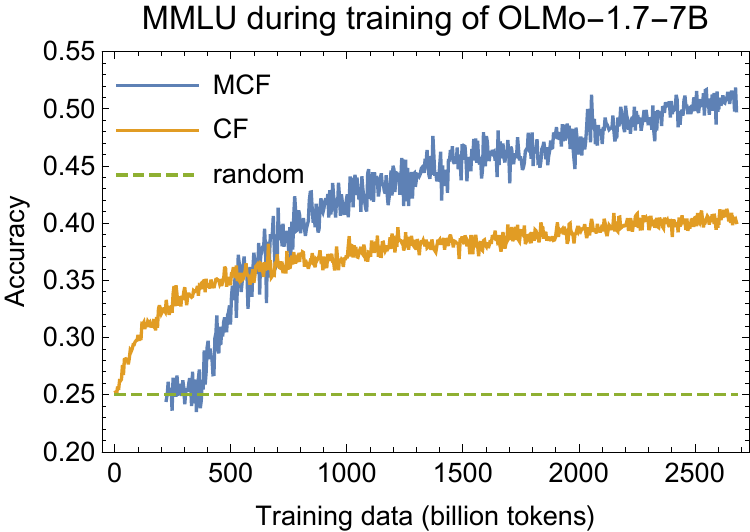}
\captionof{figure}{Performance on \task{MMLU} validation set during the training of OLMo-7B-0424 model. During early training, there is good signal from \rc{} while \mc{} is random. Around 400B tokens, the model starts gaining the ability on the \mc{} format, becoming a stronger signal than \rc{}.}
\label{fig:olmo-mmlu-train}
\end{figure}

We can see an explicit example of a model acquiring the ``understanding'' of \mc{} during training in Figure~\ref{fig:olmo-mmlu-train}, showing the OLMo-7B-0424 model \citep{groeneveld2024olmo, olmo-1-7} evaluated on the \task{MMLU} validation set in both \rc{} and \mc{} variations. The plot suggests that model starts learning the \mc{} task format after about 400 billion training tokens, so in early training \rc{} provides a better signal, while \mc{} is significantly better in late training where \rc{} levels off.

\begin{figure*}[th]
\centering
\includegraphics[width=\linewidth]{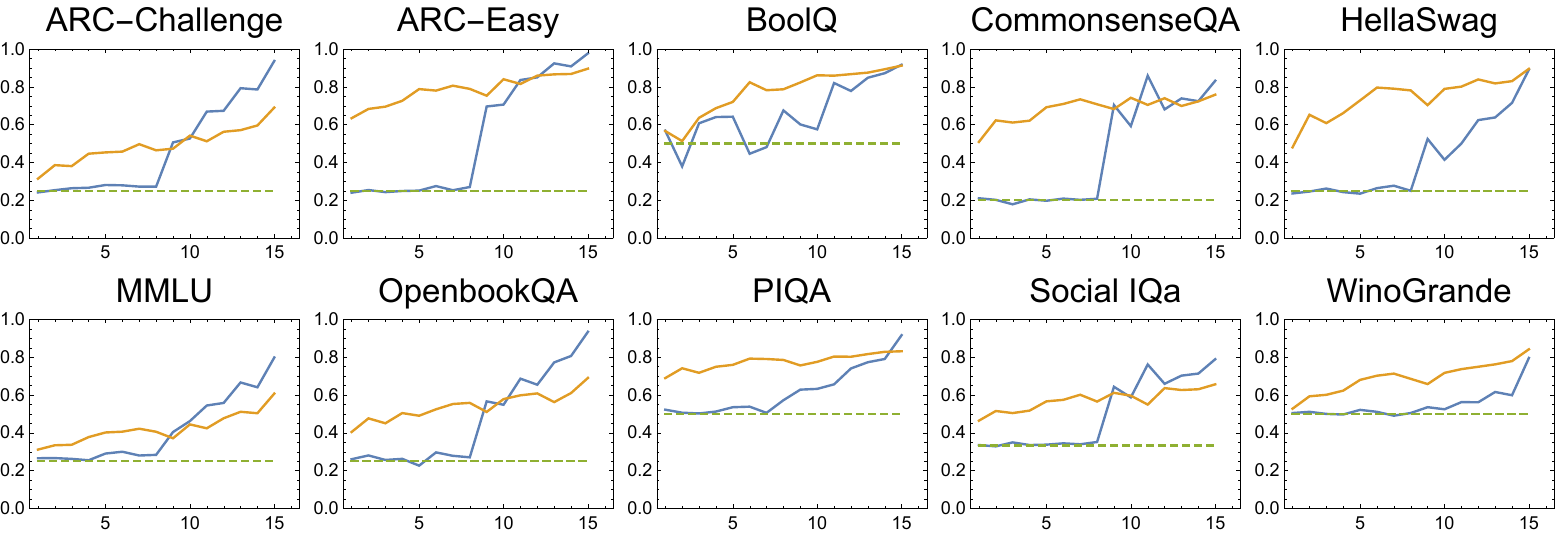}
\includegraphics[width=\columnwidth]{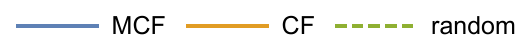}
\caption{Comparing model performance on each task for \mc{} vs \rc{}. The 15 models are ordered along the x-axis by overall performance across all 10 tasks. In general, \rc{} is needed to elicit a non-random signal from the weaker models, while stronger models can take advantage of \mc{} for a more accurate assessment.}
\label{fig:mc-rc-all}
\end{figure*}

To further study this phenomenon, we evaluate the \rc{} and \mc{} versions for each task and model.\footnote{More detailed numbers can be found in Tables~\ref{tab:appendix-task-scores1} and \ref{tab:appendix-task-scores2} in the Appendix~\ref{detailed_rc_mc_task_scores}.}  Figure~\ref{fig:mc-rc-all} shows for each task, the \mc{} and \rc{} performances for the 15 models ordered along the x-axis by overall performance on all tasks. For instance, on \task{ARC-Challenge}, we see a clear distinction where the weakest 8 models have near-random performance on the \mc{} version of the task, yet above random when using \rc{} which offers a better signal to the relative strength of models. For the stronger models, the \mc{} version clearly outscores the \rc{} version, and is a much better representation of task performance compared to the flatter trends using \rc{} (for Llama3-70B the \mc{} score is 93.7\% (6.3\% error) while the \rc{} score is just 69.0\% (31\% error), a nearly 5x difference in error rate!). 

A similar pattern can be seen across other tasks in Figure~\ref{fig:mc-rc-all}, where the stronger models
show performance using \mc{} either exceeding \rc{} (like \task{ARC-Easy}, \task{OpenbookQA}, \task{MMLU}, \task{Social IQa}, \task{CommonsenseQA}, and \task{PIQA}) or at least catching up to it (\task{HellaSwag}, \task{WinoGrande}, \task{BoolQ}).\footnote{Appendix~\ref{rc-tasks-appendix} provides further discussion.} 



In \textbf{\evalstandard{}}, we standardize to evaluate each model using both the \mc{} and \rc{} formulations, and the best performing one is used. This allows for meaningful comparison of task evaluation numbers over a range of models, from the smaller, weaker base models which can only deal with the \rc{} (where \mc{} scores hovering around random baseline), to the stronger models which can report more accurate performance using the \mc{} (where \rc{} provides less clear signal).

\subsection{Other implementation details}
\label{sec:other-specifications}
There are other important details that go into a fully specified evaluation result, and we enumerate the choices made in \evalstandard{} here:

\begin{itemize}
\itemsep0em
\item For \task{MMLU}: use macro average (over 57 tasks) rather than micro average (over 14042 instances), following \citet{llama3modelcard}. This better represents the diversity of fields in the dataset, although in practice it does not generally make a big difference (see Figure~\ref{tab:appendix-mmlu-macro-micro}).
\item When a model requires it, make sure to add the appropriate \prompt{<bos>} token at start of prompt (e.g., Gemma \citep{gemmateam2024gemma}).
\item When using the ``character'' normalization for \rc{}, include the leading space in the calculation of answer length.
\item Restrict all inputs (with completions) to 2048 tokens for consistency across models.{\footnote{For current tasks this is only exhausted for a few \task{MMLU} instances.}}
\item Use the default model precision when evaluating (i.e., avoid options like {\tt load\_in\_8bit} unless it produces identical results).
\item \evalstandard{} uses the standard approach of two newlines to separate each in-context example.
\item Other than the original instruction line for \task{MMLU} \cite{hendryckstest2021}, we do not add any extra instructions. This is in view of previous work finding the subject information from instructions makes little changes to model ranking \citep{alzahrani2024benchmarks}, and to reduce additional sources of variation in the prompt. 
\end{itemize}

Note that computational details, like batch size and type/state of GPU, can affect floating point operations such that answer choice decisions can flip if they are very close. This is hard to avoid unless one considers ``ties'' when answers are sufficiently close in confidence, we leave that for future consideration. A reference implementation of \evalstandard{} is released at \evalstandardurl{} under the Apache 2.0 license.

\section{\evalstandard{}: Summary and results}
\label{sec:reconcile-differences}


\evalstandard{} includes the following elements, justified in detail above:
\begin{itemize}
\itemsep0em
\item Use test set when available, otherwise validation. Sample 1000 instances if more than 1500 (Section~\ref{sec:varied-dataset-use})
\item Use specified, exact prompt format (Section~\ref{sec:varied-dataset-use})
\item Use fixed, curated 5-shot examples (Section~\ref{section-varied-shots})
\item Use prescribed probability normalization for \rc{} (Section ~\ref{different_normalization})
\item Evaluate with both \mc{} and \rc{}, use the best result (Section~\ref{sec:varied-format-mc-vs-rc})
\item Follow recommendations for all other evaluation details (in Section~\ref{sec:other-specifications})
\end{itemize}

Table~\ref{score-table-main} reports the overall, fully reproducible, \evalstandard{}
scores for the 15 models across the benchmarks. An extended table with a total of 40 models is shown in Table~\ref{score-table-main-extended} (Appendix~\ref{appendix-extended-results}).

\begin{table*}
\setlength\tabcolsep{2.5pt} 
  \centering
\begin{small}
\begin{tabular}{lccccccccccc}
\toprule
\footnotesize{\bf{model}} & \footnotesize{\bf{ARC\_C}} & \footnotesize{\bf{ARC\_E}} & \footnotesize{\bf{BoolQ}} & \footnotesize{\bf{CSQA}} & \footnotesize{\bf{HSwag}} & \footnotesize{\bf{MMLU}} & \footnotesize{\bf{OBQA}} & \footnotesize{\bf{PIQA}} & \footnotesize{\bf{SIQA}} & \footnotesize{\bf{WinoG}} & \footnotesize{\bf{average}}\\
\midrule
Pythia-1B & \ccell{31.4} & \ccell{63.4} & $\ccell{56.8}^\dagger$ & \ccell{50.9} & \ccell{48.0} & \ccell{31.1} & \ccell{40.4} & \ccell{68.9} & \ccell{46.4} & \ccell{52.7} & \ccell{49.0}\\
OLMo-1B & \ccell{38.6} & \ccell{68.3} & \ccell{51.3} & \ccell{62.2} & \ccell{65.2} & \ccell{33.4} & \ccell{47.6} & \ccell{74.1} & \ccell{51.5} & \ccell{59.3} & \ccell{55.1}\\
TinyLlama-1.1B & \ccell{38.1} & \ccell{69.5} & \ccell{63.6} & \ccell{61.1} & \ccell{60.8} & \ccell{33.6} & \ccell{45.0} & \ccell{71.7} & \ccell{50.4} & \ccell{60.1} & \ccell{55.4}\\
Pythia-6.7B & \ccell{44.6} & \ccell{72.6} & \ccell{68.7} & \ccell{62.1} & \ccell{66.1} & \ccell{37.7} & \ccell{50.4} & \ccell{74.9} & \ccell{51.7} & \ccell{62.3} & \ccell{59.1}\\
RPJ-INCITE-7B & \ccell{45.3} & \ccell{78.8} & \ccell{72.0} & \ccell{69.2} & \ccell{72.8} & \ccell{40.1} & \ccell{49.0} & \ccell{75.9} & \ccell{56.6} & \ccell{68.0} & \ccell{62.8}\\
StableLM2-1.6B & $\ccell{50.6}^\dagger$ & \ccell{75.3} & \ccell{82.3} & $\ccell{70.4}^\dagger$ & \ccell{70.3} & $\ccell{40.4}^\dagger$ & $\ccell{56.6}^\dagger$ & \ccell{75.6} & $\ccell{64.3}^\dagger$ & \ccell{65.7} & \ccell{65.1}\\
OLMo-7B & \ccell{46.4} & \ccell{78.9} & \ccell{78.7} & \ccell{70.8} & \ccell{78.1} & \ccell{40.5} & \ccell{55.8} & \ccell{78.5} & \ccell{56.5} & \ccell{68.5} & \ccell{65.3}\\
MPT-7b & \ccell{45.7} & \ccell{78.0} & \ccell{82.4} & \ccell{70.9} & \ccell{79.6} & \ccell{40.6} & \ccell{52.4} & \ccell{79.2} & \ccell{57.4} & \ccell{70.2} & \ccell{65.6}\\
Falcon-7B & \ccell{49.7} & \ccell{80.6} & \ccell{78.2} & \ccell{73.4} & \ccell{79.0} & \ccell{42.1} & \ccell{55.2} & \ccell{79.0} & \ccell{60.1} & \ccell{71.3} & \ccell{66.9}\\
Llama2-7B & \ccell{54.2} & \ccell{84.0} & \ccell{86.1} & \ccell{74.2} & \ccell{78.9} & $\ccell{46.2}^\dagger$ & \ccell{57.8} & \ccell{77.5} & \ccell{59.6} & \ccell{71.7} & \ccell{69.0}\\
Llama2-13B & $\ccell{67.3}^\dagger$ & \ccell{85.9} & \ccell{86.7} & \ccell{74.0} & \ccell{83.9} & $\ccell{55.8}^\dagger$ & $\ccell{65.4}^\dagger$ & \ccell{80.2} & $\ccell{65.9}^\dagger$ & \ccell{74.9} & \ccell{74.0}\\
OLMo-7B-0424 & $\ccell{66.9}^\dagger$ & $\ccell{83.6}^\dagger$ & \ccell{85.9} & $\ccell{85.8}^\dagger$ & \ccell{80.1} & $\ccell{54.4}^\dagger$ & $\ccell{68.6}^\dagger$ & \ccell{80.3} & $\ccell{76.1}^\dagger$ & \ccell{73.6} & \ccell{75.5}\\
Llama3-8B & $\ccell{79.3}^\dagger$ & $\ccell{92.4}^\dagger$ & \ccell{87.5} & $\ccell{73.9}^\dagger$ & \ccell{81.8} & $\ccell{66.6}^\dagger$ & $\ccell{77.2}^\dagger$ & \ccell{81.6} & $\ccell{70.2}^\dagger$ & \ccell{76.2} & \ccell{78.7}\\
Mistral-7B-v0.1 & $\ccell{78.6}^\dagger$ & $\ccell{90.8}^\dagger$ & \ccell{89.3} & $\ccell{72.4}^\dagger$ & \ccell{83.0} & $\ccell{64.0}^\dagger$ & $\ccell{80.6}^\dagger$ & \ccell{82.8} & $\ccell{71.3}^\dagger$ & \ccell{77.9} & \ccell{79.1}\\
Llama3-70B & $\ccell{93.7}^\dagger$ & $\ccell{97.7}^\dagger$ & $\ccell{91.7}^\dagger$ & $\ccell{83.2}^\dagger$ & \ccell{89.5} & $\ccell{79.8}^\dagger$ & $\ccell{93.4}^\dagger$ & $\ccell{91.6}^\dagger$ & $\ccell{78.9}^\dagger$ & \ccell{84.1} & \ccell{88.4}\\
\bottomrule
\end{tabular}
\end{small}
\setlength\tabcolsep{6pt} 
  \caption{Reproducible performance scores across models and tasks using \evalstandard{}, providing robust, meaningful comparisons across a wide range of models and tasks. $^\dagger$ indicates use of the \mc\ score. 
  }
  \label{score-table-main}
\vspace{-2mm}
\end{table*}

\section{Related work}
With model releases, performance on popular datasets is used to gauge the progress achieved e.g., \citet{openai2024gpt4} showing superhuman performance on benchmarks like MMLU.
Such evaluation also guides community efforts towards understanding and sharing findings on what it takes to build a strong model (\citet{touvron2023llama,touvron2023llama2,biderman2023pythia,almazrouei2023falcon,mosaic-mpt,jiang2023mistral,gemmateam2024gemma,groeneveld2024olmo} \textit{inter alia}). 
However, given a model and a dataset, even for the frequently used datasets, there are varied practices in how accuracy on them is measured. Various work has shown that model evaluations are vulnerable to differences such as option position changes in multiple-choice questions \citep{zheng2024large, li2024multiplechoice}, choice symbols, re-ordering of answer options, changing number of answer options \citep{wang2024answers}, and task formulation \citep{alzahrani2024benchmarks, robinson2023leveraging, khatun2024study, wiegreffe-etal-2023-increasing}. Even minor formatting changes can cause large, generally arbitrary, score variations \citep{sclar2023quantifying}.

The Holistic Evaluation of Language Models (HELM) benchmark \citep{liang2023holistic}, the Hugging Face Open LLM Leaderboard \citep{open-llm-leaderboard}, Mosaic Eval Gauntlet \citep{mosaic-eval-gauntlet}, Eleuther LM Evaluation Harness \citep{eval-harness, biderman2024lessons}, and Unitxt \citep{bandel-etal-2024-unitxt} present efforts toward greater transparency and reproducibility of LLM evaluations. 
These frameworks generally describe and provide support for various task setups, presenting them as open choices to researchers and users. When specific default setups are given, the rationale is not always documented and thus not followed by others in subsequent work (see Tables~\ref{score-table-variations-arc-c} and ~\ref{score-table-variations-obqa}).

\section{Discussion}

By identifying and reviewing common evaluation practices in the community, and performing experiments to resolve open questions, we present \evalstandard{} -- an open, documented, reproducible, and practical evaluation standard. OLMES provides justified recommendations on decisions such as how to format dataset instances, the choice of in-context examples, task formulation, probability normalization, as well as other implementation details. The goal is for \evalstandard{} to be a useful guide for model developers to obtain signals as to whether their model is on track during training, and to compare final powerful base models. The practical choices encourage evaluations without unnecessary computation resources. The reproducible nature means that any evaluation done using \evalstandard{} can be directly compared to existing \evalstandard{} evaluations. We also document the rationales behind the choices made, guiding the community toward more justified evaluation practices. \evalstandard{} can be applied to current leaderboards and evaluation code bases to unify evaluation practices in the field. 


\textbf{Future work and limitations.}  Future work includes adding more tasks to \evalstandard{}, covering tasks beyond MCQA such as generative tasks and chain-of-thought prompting. 
This will include standardizing how answers are extracted for evaluation, and for chat models how to split the prompt into messages. We welcome the community to contribute to \evalstandard{}, extending the principles of \evalstandard{} to new tasks. 

\evalstandard{} is a step towards standardizing LLM evaluations, ready to be incorporated into evaluation code bases for broad usage. \evalstandard{} facilitates robust and simplified comparisons of model performances, both for researchers during model training and development, and for developers in choosing models to build upon. 

\section*{Acknowledgments}
In creating this evaluation standard, \evalstandard{}, we build on top of the various previous efforts on language model evaluation in the community -- including previous work on language model evaluation standardization, the many open research reports disclosing how evaluation on LLMs have been done, and the datasets that made \evalstandard{} possible, which we explicitly cite and acknowledge in our paper.


\section*{Limitations}
The current version of \evalstandard{} is focused on providing guidance useful for LLM evaluation during the training stage and for comparing final base models, which provides important insights into the potential of such models before further tuning (e.g., instruction-tuning, safety-tuning). Interesting directions for future work include looking into evaluations targeted at accessing the effectiveness of various kinds of model tuning, as well as evaluation for multi-modal models.

This paper focuses on design choices in evaluating language models with multiple-choice tasks. While the suite of multiple-choice tasks used in this work includes questions on science, various types of commonsense, factual knowledge, and covers a range of topics (MMLU alone covers 57 subjects), of varying difficulty, an important future direction would be to apply the same principles in \evalstandard{} (e.g., prompt formatting, curated few-shot examples) to generative tasks and chain-of-thought prompting. 

While the recommendations in \evalstandard{} are well-considered, justified and practical, they do not cover all plausible variants of presenting a task. See Appendix~\ref{appendix-faq} for further discussion, showing how performance measured using OLMES is stable and consistent when subject to small changes in prompt wording or the selection of few-shot examples, within the general recommendations. Larger differences would be expected when diverging from OLMES recommendations such as by using unnatural prompts e.g., using rare symbols as answer labels, or randomly sampled few-shot examples which could run into skewed label distribution covered in few-shot examples or include noisy examples from train sets. We leave evaluating the robustness of models under adversarial setups as a topic for future work.

\section*{Ethical considerations}
This study involves the use of large-scale language models. We only use their outputs to obtain their answers to questions in commonly used multiple-choice datasets, therefore we do not foresee any ethical issues with their use for the research presented in this work.

\bibliography{anthology,custom}

\appendix
\onecolumn

\section{FAQs}
\label{appendix-faq}
 \begin{itemize}
     \item [\textbf{Q:}] \textbf{Are there existing established protocols that differ from the proposed ones? Will they cause resistance from the research community in adopting the OLMES standard? }
     \item [] The lack of “established protocols” is precisely the issue that our work addresses. As discussed in our Introduction and Related work sections, existing efforts either present \textbf{different variations as open choices to users} or are \textbf{underspecified} or \textbf{under-documented (e.g., the rationale behind default choice is unexplained)} and thus not followed by others in subsequent work. Therefore, there is no existing established protocol in the community – researchers across the field use different evaluation setups, with reasons behind their choices left unexplained, leading to different results and conclusions. We illustrate this in Table~\ref{score-table-variations-arc-c}, with Table~\ref{score-table-variations-obqa} showing an extended version to include score variations across different references on \task{OpenbookQA} in addition to \task{ARC-Challenge}.

    This is the main motivation for the standard, to reconcile the differences in practices so that scores reported in papers can be meaningfully interpreted and compared (as we show, a statement like ``score on 25-shot ARC Challenge'' is woefully underspecified, whereas ``score on ARC Challenge using OLMES'' is a well-defined number without any ambiguity). We also justify each decision we make so that the community can, for the first time, appreciate the rationale behind the setups and thus encourage broad adoption.
    
     \item [\textbf{Q:}] \textbf{What is novel about OLMES?}
     \item [] Building on the many existing works that introduce new methodologies (e.g., new way of prompting, probability normalization, etc), OLMES is the first work of its kind to provide a \textbf{completely open, practical, reproducible, and documented evaluation standard with justified choices} so that results across research work can be meaningfully compared. This fills an important gap in current research on LLMs – the adoption of OLMES by model developers and other researchers will help unify evaluation practices in the field for the first time, significantly shifting current research paradigms.

     \item [\textbf{Q:}] \textbf{Why is \evalstandard{} more principled than trying a range of settings?}
     \item [] Rather than having to train a model from scratch to discover patterns in task formulation, run the different settings to choose a normalization scheme, or delve into the same literature again to study the variants, the community can now directly build upon the various choices in OLMES.

    We hope to guide the community towards more well-documented and justifiable chosen evaluation settings like OLMES without having to go through trying a mix of less informed choices (which we argue should be avoided altogether). Through extensive literature review and experimentation, we observe  that some settings provide better signals than others, and document them in this work to guide the community to use them, a few examples include:
    
    \begin{itemize}
    \item CF gives a clearer signal early in training, which is helpful for developers to cheaply make modeling decisions. On the other hand,  MCF is a better indicator of performance later on. CF often works better for weaker models while MCF is at random, and MCF is a better representation of task performance for stronger models.
    \item Few-shot prompting is an effective and universal way to convey a task to an LLM (more stable learning curve than 0-shot) but going beyond 5 shots generally does not provide meaningful differences in scores.
    \item For probability normalization, in \task{BoolQ} the only answer choices are “yes” or “no” which are single tokens, therefore no length normalization is needed. Even if for some models, the “character” normalization has slightly better performance on \task{BoolQ} (see Table 9), one should note that this is an accidental side effect of “yes” having one more character than “no” and indeed a normalization which changes the probability of “yes” vs “no” simply because the “no” token has fewer characters seem problematic. 
    \end{itemize}
    Not only are recommendations in OLMES backed by both existing literature and new experimental results, there is also little difference between the OLMES recommendation and the empirically best (“oracle”) normalization for each task and model, see “diff oracle” column. We argue that adopting such an approach is a better practice than blindly optimizing for the best performance e.g., problematically using “character” normalization for \task{BoolQ}.

     \item [\textbf{Q:}] \textbf{Including a broader range of datasets?}
     \item [] The focus on multiple-choice datasets in OLMES is motivated by their frequent use in evaluating base LLMs, where the evaluation seems straightforward (did the model predict the right answer?) but in practice, a statement like ``model X scores Y on ARC Challenge'' is generally uninterpretable (with unspecified details and cannot be meaningfully compared across references) without a clear evaluation standard like OLMES. In this paper, we focused on datasets chosen to provide guidance useful for the training stage and evaluation of base models, which provides important insights into the potential of such models before further tuning (e.g., instruction-tuning). 

Note that the fundamental principles of OLMES as introduced, generalize to any dataset of interest. Rather than viewing what we have illustrated in our paper as a fixed set, our goal is to use that as an illustration and empower researchers to move towards reproducible evaluation by applying OLMES to any dataset of interest suited for their own work.

    \item [\textbf{Q:}] \textbf{Evaluating on more models? What are some valuable insights from extended experiments?
}
     \item [] We provide additional results in Appendix~\ref{appendix-extended-results}, Table~\ref{score-table-main-extended} with additional models. Evaluating different models using OLMES provides valuable insights for understanding LLMs and model development. For instance, within each batch of model release by developers, models of bigger size perform better than smaller ones (see average scores of Pythia-6.7B outperforms Pythia-1B, OLMo-7B outperforms OLMo-1B, Llama2-13B outperforms Llama2-7B, Gemma2-9B outperforms Gemma2-2B). However, size is not the only way to get to a stronger model, evaluating on OLMES also allow the community meaningful comparison of models to witness the effect of model improvement via better training data, model architecture, as well as other improved approaches as researchers iterate on their models e.g., OLMo-7B-0424’s improvement over initial OLMo-7B; Llama3-8B’s improvement over Llama2-7B and even Llama2-13B, Mixtral-8x7B-v0.1 outperforming the Mistral model, aligning with the insights reported in these model releases documenting their improved recipes and innovations for better models. Further, OLMES also gives meaningful comparison of model performance as researchers experiment to reduce computational costs, e.g., our results align with the original DeepSeekMoE paper where they “scale up DeepSeekMoE to 16B parameters and show that it achieves comparable performance with LLaMA2 7B, with only about 40\% of computations”. All these underscore the applicability and value of OLMES in supporting unified evaluation as the field progress towards better models, as an open, well-documented, practical and reproducible evaluation standard. We make all prompts, examples, and code used for OLMES openly available, and encourage researchers to try it for any model of their interest be it one they are studying or building.

         \item [\textbf{Q:}] \textbf{How does OLMES stay relevant in the rapid evolution of AI and LLMs?}
     \item [] We have been continuously looking out for new LLMs and evaluating them using OLMES, showing that the same guiding principles still apply as best practices providing a systematic, comparable approach. See extended evaluation results in Table~\ref{score-table-main-extended}.

    As the field moves forward, we look forward to applying the principles of OLMES to more benchmarks and evaluating newer models using OLMES. While we are working on extending OLMES, we do not anticipate revisions to the currently established recommendations in OLMES any time soon as the guiding principles are built on top of a rich literature of existing work over the years and will likely remain relevant in the community for a while in the near future.

    \item [\textbf{Q:}] \textbf{Why not use prompting techniques such as CoT or self-reflection?}
     \item [] While these prompting strategies have shown to be useful for instruction-tuned models, they tend to be much less effective on base models, which is a focus of this work. E.g., some experiments we performed with MMLU showed that various CoT prompts (both zero-shot and few-shot) have a positive boost on instruction-tuned models (like Llama-3.1-8B-Instruct), but tend to lower the scores a bit for base models (like Llama-3.1-8B).

    \item [\textbf{Q:}] \textbf{How do you ensure that formatting settings are fair to all models?}
     \item [] Supported by reviewing common evaluation practices in the community and empirical evidence across a wide range of models, the recommendations we make are at least as reasonable and fair as the myriad of settings that have been used in the literature. 

If a model is peculiar in any specific way (e.g., only able to do multiple choice questions with one type of answer label like “1.” or “2.”), it is not the goal of OLMES to tailor to such peculiarities as this standard is intended to be applied across a range of models and to encourage the development of models that produce reasonable outputs given any reasonable input.

\item [\textbf{Q:}] \textbf{What happens when there are minor variants to OLMES?}
     \item [] Through OLMES, we provide best practices to evaluate language models and justify our choices. Our choices are mostly aligned with common practices in LLM evaluations, but with defining standards in formatting, choice of in-context examples, probability normalizations, and task formulation. In the process, we accounted for many factors, taking into consideration the robustness of OLMES under minor variations. We discuss some of these considerations here. 

     \textbf{[Part 1] Order of presenting the options A/B/C/D:}
     
     The order of presenting the multiple-choice options A/B/C/D does not apply to CF since each answer is processed independently. For MCF it is indeed a confounder that some (especially weaker) models might highly prefer a given label (like B). The benchmarks in OLMES are generally balanced such that such a model would not be much better than random. Further, if this happens, CF would generally get a better score in such cases and OLMES would use that score in its final output. Therefore, having a setting where we use both CF (not affected by the order of options) and MCF (where the order of options may matter) makes sure the final metric will not be hugely affected by such factors. We considered applying more rigorous measures (like running all cyclic permutations of answer choices) but decided for practical reasons, the extra processing time and complexity were not worth the minor improvements in robustness (as one consideration of OLMES is also to be a practical standard that does not take unnecessarily more compute than is needed).

     \textbf{[Part 2] Minor variations in prompt wording or few-shot examples:}
     
     To address potential concerns on minor variations in prompt wording or few-shot examples, we evaluated under three additional settings, while adhering to the general principles in OLMES:

\textbf{Variant 1 (minor variation in prompting):}\\
3 changes to OLMES prompt format - (1) change the label and text separator from  “.” to  “)”, (2) insert an additional new line before the answer descriptor, (3) change the “Answer” descriptor to “Correct answer”\\
\textbf{Variant 2 (varying few-shot examples):}\\
Create a different set of curated few-shot examples by changing 3 out of the 5 in-context examples to new ones that are different from those in OLMES. In picking the new few-shot examples, the same recommendations were followed to ensure diversity in the examples and that they cover the label space.\\
\textbf{Variant 3 (minor variation in prompting + varying few-shot examples):}\\
Apply changes in both Variants 1 and 2 together.\\

We report these additional results in Table~\ref{score-table-main-vars}. Following EleutherAI in calculating standard error (\url{https://github.com/EleutherAI/lm-evaluation-harness/blob/ebe7226ebfb8d11a9fb8d6b53eb65891f895c633/lm_eval/api/metrics.py#L288}), in the additional results, we also incorporated bounds on standard error in our evaluations using OLMES (see “std err” column).  This provides a statistical bound on the degree of variation in reported numbers and illustrates that while any performance metric should be interpreted to have slight variants (e.g., < 2.5\%), the scenario where a model underperforms significantly due to minor variants is unlikely statistically.

\begin{table*}
  \centering
\begin{tabular}{lccccccc}
\toprule
\bf{model} & \bf{ARC\_E orig} & \bf{var 1} & \bf{var 2} & \bf{var 3} & \bf{avg} & \bf{diff} & \bf{std err}\\
\midrule
Pythia-1B & 63.4 & 63.3 & 62.7 & 62.7 & 63.0 & 0.4 & 1.5\\
Llama2-7B & 84.0 & 84.4 & 83.4 & 84.4 & 84.0 & 0.0 & 1.2\\
DeepSeek-7B & 80.6 & 80.9 & 80.5 & 80.4 & 80.6 & 0.0 & 1.3\\
Gemma2-2B & $84.3^\dagger$ & $83.2^\dagger$ & $83.9^\dagger$ & $82.8^\dagger$ & 83.5 & 0.8 & 1.2\\
Llama3-8B & $92.4^\dagger$ & $92.5^\dagger$ & $92.3^\dagger$ & $93.1^\dagger$ & 92.6 & 0.2 & 0.8\\
\bottomrule
\toprule
\bf{model} & \bf{OBQA orig} & \bf{var 1} & \bf{var 2} & \bf{var 3} & \bf{avg} & \bf{diff} & \bf{std err}\\
\midrule
Pythia-1B & 40.4 & 38.6 & 39.4 & 37.6 & 39.0 & 1.4 & 2.2\\
Llama2-7B & 57.8 & 57.2 & 55.2 & 57.4 & 56.9 & 0.9 & 2.2\\
DeepSeek-7B & $62.2^\dagger$ & $61.0^\dagger$ & $61.6^\dagger$ & $63.2^\dagger$ & 62.0 & 0.2 & 2.2\\
Gemma2-2B & $68.8^\dagger$ & $67.2^\dagger$ & $68.8^\dagger$ & $67.0^\dagger$ & 68.0 & 0.8 & 2.1\\
Llama3-8B & $77.2^\dagger$ & $76.8^\dagger$ & $78.8^\dagger$ & $77.8^\dagger$ & 77.7 & 0.5 & 1.9\\
\bottomrule
\toprule
\bf{model} & \bf{PIQA orig} & \bf{var 1} & \bf{var 2} & \bf{var 3} & \bf{avg} & \bf{diff} & \bf{std err}\\
\midrule
Pythia-1B & 68.9 & 69.2 & 69.2 & 69.3 & 69.1 & 0.2 & 1.5\\
Llama2-7B & 77.5 & 77.2 & 77.7 & 77.8 & 77.5 & 0.0 & 1.3\\
DeepSeek-7B & 79.3 & 78.8 & 80.9 & 80.6 & 79.9 & 0.6 & 1.3\\
Gemma2-2B & 78.5 & 77.8 & 79.3 & 78.5 & 78.5 & 0.0 & 1.3\\
Llama3-8B & 81.6 & 80.7 & 82.4 & 82.8 & 81.9 & 0.3 & 1.2\\
\bottomrule
\end{tabular}
\setlength\tabcolsep{6pt} 
  \caption{Extended results comparing using OLMES (orig) and when the setting is subjected to minor variations in prompt wording (var1), few-shot examples (var2), or both (var3). $\dagger$ indicates the use of MCF. The ``avg'' score obtained via averaging orig, var1, var2, and var3 results is often within 1\% of that obtained by the original OLMES setup (orig). We report the observed differences between averaging the 4 setups (``avg'') and directly using OLMES (orig) in the ``diff'' column, illustrating the minor differences (often <1\%) do not justify the 4 times more compute needed, against the ``practical'' consideration in OLMES. 
  }
  \label{score-table-main-vars}
\end{table*}

The additional results show that differences in performance between averaging variations vs. using the OLMES setup directly were generally minimal, typically less than 1 percent (the largest difference seen is 1.4\%). This suggests that performance measured using OLMES is quite stable and consistent when subject to small changes in prompt wording or the selection of few-shot examples, within the general recommendations. Note that these variants still format the instances in natural ways and are slight modifications of the original settings of OLMES, still adhering to the general principles such as instance formatting that clarifies the task in a natural way and choice of in-context examples to cover a range of examples and different answer labels. Larger differences would be expected when diverging from OLMES recommendations such as by using unnatural prompts e.g., using rare symbols as answer labels, or randomly sampled few-shot examples which could run into skewed label distribution covered in few-shot examples or include noisy examples from train sets. We observe that current successful language models are generally robust to the OLMES evaluation standard. OLMES has been informed by prior efforts like HELM and Eleuther LM Evaluation Harness, therefore the prompts are designed to be natural, and suitable for evaluating language models.

\end{itemize}

\section{Detailed \rc{} and \mc{} task scores}
\label{detailed_rc_mc_task_scores}
Tables~\ref{tab:appendix-task-scores1} and ~\ref{tab:appendix-task-scores2} present detailed scores across all tasks, with both \mc{} and \rc{} results (using the \evalstandard{} recommendations for \rc{} normalization).

\section{Further details on variations}
In this appendix we discuss further details on how LLM evaluations can vary and the choices made in \evalstandard{}.

\subsection{Task formulation details}
\label{task-formulation-details-appendix}
LLM evaluations started out using the \rc{} approach for many tasks \citep{GPT3NEURIPS2020, du2022glam, smith2022megatron530b, chowdhery2022palm, jurassic}, which is a more reasonable option for weaker models that struggle with the more natural \mc{} \citep{khatun2024study}. The task formulation only very recently and gradually switched to the \mc{} approach when it became clear that the model could utilize it, producing higher scores \citep{robinson2023leveraging, openai2024gpt4,llama3modelcard}.

The HELM study \citep{liang2023holistic} included comparisons between the \mc{} (``joint'') and \rc{} (``separate'') approaches,
finding that certain models can really benefit from the \mc{} approach, although among the models in the original study it was really only the Anthropic-LM v4-s3 (52B) model which could take full advantage of it.

\subsection{\rc{} normalization details}
\label{normalization-details-appendix}

Tables ~\ref{tab:appendix-rc-norm-1}, ~\ref{tab:appendix-rc-norm-2} and ~\ref{tab:appendix-rc-norm-3} show detailed comparisons of \rc{} normalization on different models, for the various tasks.

Unlike in \mc{}, where the evaluation metric involves just scoring the log-likelihood corresponding to the answer choice label (i.e., A/B/C/...), there is a choice of log-likelihood normalization (``none'', ``per token'', ``per character'' or ``pmi'') for \rc{} as detailed in Section \ref{different_normalization}. 

When evaluating the GPT-3 model \citep{GPT3NEURIPS2020}, they worked around this issue by normalizing the log-probability by the number of tokens in the answer (similar to how loss is computed during training). They also noted that for a few datasets, it worked markedly better to instead ``normalize'' by dividing by LLM probability of the same answer string without the presence of the question (usually by just having a generic prefix like {\tt \frenchspacing "Answer: <answer\_string>"}). This can be considered a form of pointwise-mutual-information (PMI) and was explored further in other works \citep{holtzman-etal-2021-surface}.

The Eleuther LM Evaluation Harness \citep{eval-harness, biderman2024lessons} and some subsequent evaluations (e.g., the Llama models \citep{touvron2023llama}) have also used ``per answer character'' normalization, using the argumentation \citep{eleuther-normalization, biderman2024lessons}, that normalizing per token is problematic since it depends on the tokenizer. Since the purpose of the normalization is simply to rank the answer choices within themselves (keeping model and tokenizer fixed), this does not seem like a relevant argument, and indeed a normalization which changes the probability of ``yes'' vs ``no'' simply because the ``no'' token has fewer characters seem problematic. In practice, for tasks where answers are either relatively long or similar in length, there are minor differences between these two length normalizations.

The HELM study \citep{liang2023holistic} included comparisons between these normalization approaches for a number of tasks and models (using the terms ``separate'' and ``separate calibrated'' for ``token'' and ``pmi'' respectively), eventually settling on a default choice for each, not unlike the choices in the GPT-3 report \citep{GPT3NEURIPS2020}. The Eleuther LM Evaluation Harness generally reports two metrics for each multiple-choice task: {\tt acc}  (using the ``none'' normalization) and {\tt acc\_norm} (using the ``character'' normalization).

\begin{table*}
  \centering

\setlength\tabcolsep{2.5pt} 
\begin{tabular}{lcccccccccccc}
\toprule
 & \multicolumn{2}{c}{\bf{ARC\_C}} & \multicolumn{2}{c}{\bf{ARC\_E}} & \multicolumn{2}{c}{\bf{BoolQ}} & \multicolumn{2}{c}{\bf{CSQA}} & \multicolumn{2}{c}{\bf{HSwag}} & \multicolumn{2}{c}{\bf{MMLU}}\\
model & \mc & \rc & \mc & \rc & \mc & \rc & \mc & \rc & \mc & \rc & \mc & \rc\\
\midrule
Pythia-1B & \ccell{24.1} & \ccell{31.4} & \ccell{24.0} & \ccell{63.4} & \ccell{56.8} & \ccell{56.6} & \ccell{21.0} & \ccell{50.9} & \ccell{23.6} & \ccell{48.0} & \ccell{26.5} & \ccell{31.1}\\
OLMo-1B & \ccell{25.3} & \ccell{38.6} & \ccell{25.4} & \ccell{68.3} & \ccell{37.9} & \ccell{51.3} & \ccell{20.2} & \ccell{62.2} & \ccell{24.6} & \ccell{65.2} & \ccell{26.6} & \ccell{33.4}\\
TinyLlama-1.1B & \ccell{26.4} & \ccell{38.1} & \ccell{24.3} & \ccell{69.5} & \ccell{60.7} & \ccell{63.6} & \ccell{17.9} & \ccell{61.1} & \ccell{26.2} & \ccell{60.8} & \ccell{26.2} & \ccell{33.6}\\
Pythia-6.7B & \ccell{26.6} & \ccell{44.6} & \ccell{24.9} & \ccell{72.6} & \ccell{64.0} & \ccell{68.7} & \ccell{20.5} & \ccell{62.1} & \ccell{24.3} & \ccell{66.1} & \ccell{25.4} & \ccell{37.7}\\
RPJ-INCITE-7B & \ccell{28.1} & \ccell{45.3} & \ccell{25.1} & \ccell{78.8} & \ccell{64.2} & \ccell{72.0} & \ccell{19.7} & \ccell{69.2} & \ccell{23.5} & \ccell{72.8} & \ccell{29.0} & \ccell{40.1}\\
StableLM2-1.6B & \ccell{50.6} & \ccell{47.3} & \ccell{69.6} & \ccell{75.3} & \ccell{60.1} & \ccell{82.3} & \ccell{70.4} & \ccell{68.2} & \ccell{52.4} & \ccell{70.3} & \ccell{40.4} & \ccell{37.1}\\
OLMo-7B & \ccell{27.2} & \ccell{46.4} & \ccell{27.0} & \ccell{78.9} & \ccell{67.5} & \ccell{78.7} & \ccell{20.8} & \ccell{70.8} & \ccell{25.0} & \ccell{78.1} & \ccell{28.3} & \ccell{40.5}\\
MPT-7b & \ccell{27.9} & \ccell{45.7} & \ccell{27.5} & \ccell{78.0} & \ccell{44.6} & \ccell{82.4} & \ccell{20.9} & \ccell{70.9} & \ccell{26.4} & \ccell{79.6} & \ccell{30.0} & \ccell{40.6}\\
Falcon-7B & \ccell{27.2} & \ccell{49.7} & \ccell{25.3} & \ccell{80.6} & \ccell{48.1} & \ccell{78.2} & \ccell{20.2} & \ccell{73.4} & \ccell{27.7} & \ccell{79.0} & \ccell{28.0} & \ccell{42.1}\\
Llama2-7B & \ccell{52.6} & \ccell{54.2} & \ccell{70.6} & \ccell{84.0} & \ccell{57.5} & \ccell{86.1} & \ccell{59.2} & \ccell{74.2} & \ccell{41.4} & \ccell{78.9} & \ccell{46.2} & \ccell{44.4}\\
Llama2-13B & \ccell{67.3} & \ccell{56.2} & \ccell{85.0} & \ccell{85.9} & \ccell{77.8} & \ccell{86.7} & \ccell{68.1} & \ccell{74.0} & \ccell{62.4} & \ccell{83.9} & \ccell{55.8} & \ccell{47.6}\\
OLMo-7B-0424 & \ccell{66.9} & \ccell{51.2} & \ccell{83.6} & \ccell{81.5} & \ccell{82.0} & \ccell{85.9} & \ccell{85.8} & \ccell{70.4} & \ccell{50.0} & \ccell{80.1} & \ccell{54.4} & \ccell{42.4}\\
Llama3-8B & \ccell{79.3} & \ccell{57.1} & \ccell{92.4} & \ccell{86.6} & \ccell{84.8} & \ccell{87.5} & \ccell{73.9} & \ccell{69.9} & \ccell{63.8} & \ccell{81.8} & \ccell{66.6} & \ccell{51.1}\\
Mistral-7B-v0.1 & \ccell{78.6} & \ccell{59.6} & \ccell{90.8} & \ccell{86.8} & \ccell{87.2} & \ccell{89.3} & \ccell{72.4} & \ccell{72.3} & \ccell{71.5} & \ccell{83.0} & \ccell{64.0} & \ccell{50.3}\\
Llama3-70B & \ccell{93.7} & \ccell{69.0} & \ccell{97.7} & \ccell{89.6} & \ccell{91.7} & \ccell{91.2} & \ccell{83.2} & \ccell{75.8} & \ccell{89.1} & \ccell{89.5} & \ccell{79.8} & \ccell{60.7}\\
\bottomrule
\end{tabular}
\setlength\tabcolsep{6pt} 
  \caption{Comparing \mc{} and \rc{} scores on each task (part 1). Weaker models at the top of the table have near-random \mc{} scores, while for stronger models at the bottom, the \mc{} score provides a better assessment than the \rc{} score.}
  \label{tab:appendix-task-scores1}
\end{table*}

\begin{table*}
  \centering

\setlength\tabcolsep{2.5pt} 
\begin{tabular}{lcccccccccccc}
\toprule
 & \multicolumn{2}{c}{\bf{OBQA}} & \multicolumn{2}{c}{\bf{PIQA}} & \multicolumn{2}{c}{\bf{SIQA}} & \multicolumn{2}{c}{\bf{WinoG}} & \multicolumn{4}{c}{\bf{average scores}}\\
model & \mc & \rc & \mc & \rc & \mc & \rc & \mc & \rc & \mc & \rc & all & max\\
\midrule
Pythia-1B & \ccell{26.0} & \ccell{40.4} & \ccell{52.2} & \ccell{68.9} & \ccell{33.5} & \ccell{46.4} & \ccell{50.4} & \ccell{52.7} & \ccell{33.8} & \ccell{49.0} & \ccell{41.4} & \ccell{49.0}\\
OLMo-1B & \ccell{28.0} & \ccell{47.6} & \ccell{50.6} & \ccell{74.1} & \ccell{32.8} & \ccell{51.5} & \ccell{51.1} & \ccell{59.3} & \ccell{32.3} & \ccell{55.1} & \ccell{43.7} & \ccell{55.1}\\
TinyLlama-1.1B & \ccell{25.6} & \ccell{45.0} & \ccell{50.2} & \ccell{71.7} & \ccell{34.9} & \ccell{50.4} & \ccell{50.0} & \ccell{60.1} & \ccell{34.2} & \ccell{55.4} & \ccell{44.8} & \ccell{55.4}\\
Pythia-6.7B & \ccell{26.2} & \ccell{50.4} & \ccell{51.2} & \ccell{74.9} & \ccell{33.5} & \ccell{51.7} & \ccell{49.6} & \ccell{62.3} & \ccell{34.6} & \ccell{59.1} & \ccell{46.9} & \ccell{59.1}\\
RPJ-INCITE-7B & \ccell{22.6} & \ccell{49.0} & \ccell{53.5} & \ccell{75.9} & \ccell{33.7} & \ccell{56.6} & \ccell{52.1} & \ccell{68.0} & \ccell{35.1} & \ccell{62.8} & \ccell{49.0} & \ccell{62.8}\\
StableLM2-1.6B & \ccell{56.6} & \ccell{51.0} & \ccell{62.8} & \ccell{75.6} & \ccell{64.3} & \ccell{61.1} & \ccell{53.5} & \ccell{65.7} & \ccell{58.1} & \ccell{63.4} & \ccell{60.7} & \ccell{65.1}\\
OLMo-7B & \ccell{27.0} & \ccell{55.8} & \ccell{57.2} & \ccell{78.5} & \ccell{35.1} & \ccell{56.5} & \ccell{50.4} & \ccell{68.5} & \ccell{36.6} & \ccell{65.3} & \ccell{50.9} & \ccell{65.3}\\
MPT-7b & \ccell{29.6} & \ccell{52.4} & \ccell{53.8} & \ccell{79.2} & \ccell{34.4} & \ccell{57.4} & \ccell{51.1} & \ccell{70.2} & \ccell{34.6} & \ccell{65.6} & \ccell{50.1} & \ccell{65.6}\\
Falcon-7B & \ccell{27.8} & \ccell{55.2} & \ccell{50.5} & \ccell{79.0} & \ccell{33.9} & \ccell{60.1} & \ccell{49.0} & \ccell{71.3} & \ccell{33.8} & \ccell{66.9} & \ccell{50.3} & \ccell{66.9}\\
Llama2-7B & \ccell{54.8} & \ccell{57.8} & \ccell{63.2} & \ccell{77.5} & \ccell{58.7} & \ccell{59.6} & \ccell{52.4} & \ccell{71.7} & \ccell{55.7} & \ccell{68.8} & \ccell{62.2} & \ccell{69.0}\\
Llama2-13B & \ccell{65.4} & \ccell{60.8} & \ccell{74.0} & \ccell{80.2} & \ccell{65.9} & \ccell{63.6} & \ccell{56.1} & \ccell{74.9} & \ccell{67.8} & \ccell{71.4} & \ccell{69.6} & \ccell{74.0}\\
OLMo-7B-0424 & \ccell{68.6} & \ccell{59.8} & \ccell{65.6} & \ccell{80.3} & \ccell{76.1} & \ccell{54.9} & \ccell{56.2} & \ccell{73.6} & \ccell{68.9} & \ccell{68.0} & \ccell{68.5} & \ccell{75.5}\\
Llama3-8B & \ccell{77.2} & \ccell{56.2} & \ccell{77.3} & \ccell{81.6} & \ccell{70.2} & \ccell{62.6} & \ccell{61.6} & \ccell{76.2} & \ccell{74.7} & \ccell{71.0} & \ccell{72.9} & \ccell{78.7}\\
Mistral-7B-v0.1 & \ccell{80.6} & \ccell{61.0} & \ccell{79.0} & \ccell{82.8} & \ccell{71.3} & \ccell{63.0} & \ccell{59.8} & \ccell{77.9} & \ccell{75.5} & \ccell{72.6} & \ccell{74.1} & \ccell{79.1}\\
Llama3-70B & \ccell{93.4} & \ccell{69.0} & \ccell{91.6} & \ccell{83.1} & \ccell{78.9} & \ccell{65.6} & \ccell{79.6} & \ccell{84.1} & \ccell{87.9} & \ccell{77.8} & \ccell{82.8} & \ccell{88.4}\\
\bottomrule
\end{tabular}
\setlength\tabcolsep{6pt} 
  \caption{Comparing \mc{} and \rc{} scores on each task (part 2), along with overall averages. The ``max'' average corresponds to the \evalstandard{} score, taking the best of \mc{} and \rc{} for each task.}
  \label{tab:appendix-task-scores2}
\end{table*}

\begin{table*}
  \centering
  \small

\setlength\tabcolsep{2pt} 
\begin{tabular}{lcccc}
\toprule
\bf{model} & \bf{\mc{}-macro} & \bf{\mc{}-micro} & \bf{\rc{}-macro} & \bf{\rc{}-micro}\\
\midrule
Pythia-6.7B & \ccell{25.4} & \ccell{25.2} & \ccell{37.7} & \ccell{37.5}\\
TinyLlama-1.1B & \ccell{26.2} & \ccell{25.7} & \ccell{33.6} & \ccell{33.5}\\
Pythia-1B & \ccell{26.5} & \ccell{26.4} & \ccell{31.1} & \ccell{31.2}\\
OLMo-1B & \ccell{26.6} & \ccell{26.3} & \ccell{33.4} & \ccell{33.6}\\
Falcon-7B & \ccell{28.0} & \ccell{27.7} & \ccell{42.1} & \ccell{41.9}\\
OLMo-7B & \ccell{28.3} & \ccell{28.3} & \ccell{40.5} & \ccell{40.7}\\
RPJ-INCITE-7B & \ccell{29.0} & \ccell{28.4} & \ccell{40.1} & \ccell{40.1}\\
MPT-7b & \ccell{30.0} & \ccell{29.3} & \ccell{40.6} & \ccell{40.6}\\
StableLM2-1.6B & \ccell{40.4} & \ccell{39.6} & \ccell{37.1} & \ccell{37.0}\\
Llama2-7B & \ccell{46.2} & \ccell{45.5} & \ccell{44.4} & \ccell{44.3}\\
OLMo-7B-0424 & \ccell{54.4} & \ccell{52.8} & \ccell{42.4} & \ccell{42.4}\\
Llama2-13B & \ccell{55.8} & \ccell{55.5} & \ccell{47.6} & \ccell{47.1}\\
Mistral-7B-v0.1 & \ccell{64.0} & \ccell{63.0} & \ccell{50.3} & \ccell{49.8}\\
Llama3-8B & \ccell{66.6} & \ccell{65.4} & \ccell{51.1} & \ccell{50.8}\\
Llama3-70B & \ccell{79.8} & \ccell{79.2} & \ccell{60.7} & \ccell{60.5}\\
\bottomrule
\end{tabular}
\setlength\tabcolsep{6pt} 
  \caption{Macro vs micro average scores on \task{MMLU}, where macro average is over the 57 tasks and micro average is over the 14042 individual questions. In general there are small differences between the two.}
  \label{tab:appendix-mmlu-macro-micro}
\end{table*}

\begin{table*}
  \centering
  \footnotesize
 
\setlength\tabcolsep{1.5pt} 
\begin{tabular}{lcccccccccccccccccc}
\toprule
 & \multicolumn{2}{c}{\bf{ARC\_C}} & \multicolumn{2}{c}{\bf{ARC\_E}} & \multicolumn{2}{c}{\bf{BoolQ}} & \multicolumn{2}{c}{\bf{CSQA}} & \multicolumn{2}{c}{\bf{HSwag}} & \multicolumn{2}{c}{\bf{MMLU}} & \multicolumn{2}{c}{\bf{OBQA}} & \multicolumn{2}{c}{\bf{PIQA}} & \multicolumn{2}{c}{\bf{SIQA}}\\
model & pmi & diff & char & diff & none & diff & pmi & diff & char & diff & char & diff & pmi & diff & char & diff & char & diff\\
\midrule
Pythia-1B & \ccell{31.4} & 0.0 & \ccell{63.4} & 0.0 & \ccell{56.6} & 4.5 & \ccell{50.9} & 0.0 & \ccell{48.0} & 0.0 & \ccell{31.1} & 1.2 & \ccell{40.4} & 0.0 & \ccell{68.9} & 1.4 & \ccell{46.4} & 0.0\\
OLMo-1B & \ccell{38.6} & 0.0 & \ccell{68.3} & 0.2 & \ccell{51.3} & 4.7 & \ccell{62.2} & 0.0 & \ccell{65.2} & 0.0 & \ccell{33.4} & 0.8 & \ccell{47.6} & 0.0 & \ccell{74.1} & 0.0 & \ccell{51.5} & 0.0\\
TinyLlama-1.1B & \ccell{38.1} & 0.0 & \ccell{69.5} & 0.0 & \ccell{63.6} & 2.2 & \ccell{61.1} & 0.0 & \ccell{60.8} & 0.0 & \ccell{33.6} & 0.9 & \ccell{45.0} & 0.0 & \ccell{71.7} & 0.6 & \ccell{50.4} & 0.0\\
Pythia-6.7B & \ccell{44.6} & 0.0 & \ccell{72.6} & 0.0 & \ccell{68.7} & 0.0 & \ccell{62.1} & 0.2 & \ccell{66.1} & 0.0 & \ccell{37.7} & 0.2 & \ccell{50.4} & 0.0 & \ccell{74.9} & 0.0 & \ccell{51.7} & 1.1\\
RPJ-INCITE-7B & \ccell{45.3} & 0.0 & \ccell{78.8} & 0.0 & \ccell{72.0} & 2.5 & \ccell{69.2} & 0.2 & \ccell{72.8} & 0.0 & \ccell{40.1} & 0.8 & \ccell{49.0} & 0.0 & \ccell{75.9} & 0.1 & \ccell{56.6} & 0.0\\
MPT-7b & \ccell{45.7} & 0.6 & \ccell{78.0} & 0.0 & \ccell{82.4} & 0.0 & \ccell{70.9} & 0.0 & \ccell{79.6} & 0.0 & \ccell{40.6} & 0.0 & \ccell{52.4} & 0.0 & \ccell{79.2} & 0.0 & \ccell{57.4} & 0.0\\
Falcon-7B & \ccell{49.7} & 0.0 & \ccell{80.6} & 0.0 & \ccell{78.2} & 0.6 & \ccell{73.4} & 0.0 & \ccell{79.0} & 0.0 & \ccell{42.1} & 0.0 & \ccell{55.2} & 0.0 & \ccell{79.0} & 0.2 & \ccell{60.1} & 0.0\\
OLMo-7B & \ccell{46.4} & 0.0 & \ccell{78.9} & 0.0 & \ccell{78.7} & 0.0 & \ccell{70.8} & 0.0 & \ccell{78.1} & 0.0 & \ccell{40.5} & 0.1 & \ccell{55.8} & 0.0 & \ccell{78.5} & 0.8 & \ccell{56.5} & 0.0\\
StableLM2-1.6B & \ccell{47.3} & 0.0 & \ccell{75.3} & 0.0 & \ccell{82.3} & 0.0 & \ccell{68.2} & 0.0 & \ccell{70.3} & 0.0 & \ccell{37.1} & 1.5 & \ccell{51.0} & 0.0 & \ccell{75.6} & 0.3 & \ccell{61.1} & 0.0\\
Llama2-7B & \ccell{54.2} & 0.0 & \ccell{84.0} & 0.0 & \ccell{86.1} & 0.0 & \ccell{74.2} & 0.0 & \ccell{78.9} & 0.0 & \ccell{44.4} & 0.4 & \ccell{57.8} & 0.0 & \ccell{77.5} & 0.2 & \ccell{59.6} & 0.0\\
OLMo-7B-0424 & \ccell{51.2} & 0.0 & \ccell{81.5} & 0.0 & \ccell{85.9} & 0.0 & \ccell{70.4} & 1.1 & \ccell{80.1} & 0.0 & \ccell{42.4} & 0.0 & \ccell{59.8} & 0.0 & \ccell{80.3} & 0.0 & \ccell{54.9} & 0.8\\
Llama2-13B & \ccell{56.2} & 0.9 & \ccell{85.9} & 0.0 & \ccell{86.7} & 1.5 & \ccell{74.0} & 0.0 & \ccell{83.9} & 0.0 & \ccell{47.6} & 0.0 & \ccell{60.8} & 0.0 & \ccell{80.2} & 0.0 & \ccell{63.6} & 0.0\\
Llama3-8B & \ccell{57.1} & 1.3 & \ccell{86.6} & 0.0 & \ccell{87.5} & 0.3 & \ccell{69.9} & 4.3 & \ccell{81.8} & 0.0 & \ccell{51.1} & 0.0 & \ccell{56.2} & 0.0 & \ccell{81.6} & 0.0 & \ccell{62.6} & 0.0\\
Mistral-7B-v0.1 & \ccell{59.6} & 0.6 & \ccell{86.8} & 0.0 & \ccell{89.3} & 0.0 & \ccell{72.3} & 2.1 & \ccell{83.0} & 0.0 & \ccell{50.3} & 0.0 & \ccell{61.0} & 0.0 & \ccell{82.8} & 0.0 & \ccell{63.0} & 0.0\\
Llama3-70B & \ccell{69.0} & 0.0 & \ccell{89.6} & 0.8 & \ccell{91.2} & 0.5 & \ccell{75.8} & 1.3 & \ccell{89.5} & 0.0 & \ccell{60.7} & 0.0 & \ccell{69.0} & 0.0 & \ccell{83.1} & 0.1 & \ccell{65.6} & 0.0\\
\bottomrule
\end{tabular}

\setlength\tabcolsep{6pt} 
  \caption{Normalization details, showing that our recommendations are not only supported by reasoning using principles behind the normalization but also close to the empirically best normalization that lets you get the highest accuracy for each model on each task (see ``diff'' columns).}
  \label{tab-rc-norm-diff}
\end{table*}

\begin{table*}
  \centering
  \small

\setlength\tabcolsep{2.5pt} 
\begin{tabular}{lccccc|ccccc|ccccc}
\toprule
 & \multicolumn{5}{c|}{\bf{ARC\_C}} & \multicolumn{5}{c|}{\bf{ARC\_E}} & \multicolumn{5}{c}{\bf{BoolQ}}\\
model & none & char & tok & pmi & best & none & char & tok & pmi & best & none & char & tok & pmi & best\\
\midrule
Pythia-1B & \ccell{26.1} & \ccell{28.4} & \ccell{29.0} & \ccell{31.4} & pmi & \ccell{61.9} & \ccell{63.4} & \ccell{60.9} & \ccell{56.5} & char & \ccell{56.6} & \ccell{61.1} & \ccell{56.6} & \ccell{41.0} & char\\
OLMo-1B & \ccell{32.9} & \ccell{34.4} & \ccell{34.7} & \ccell{38.6} & pmi & \ccell{68.5} & \ccell{68.3} & \ccell{65.8} & \ccell{60.2} & none & \ccell{51.3} & \ccell{56.0} & \ccell{51.3} & \ccell{42.3} & char\\
TinyLlama-1.1B & \ccell{31.5} & \ccell{34.1} & \ccell{32.2} & \ccell{38.1} & pmi & \ccell{68.6} & \ccell{69.5} & \ccell{64.4} & \ccell{60.4} & char & \ccell{63.6} & \ccell{65.8} & \ccell{63.6} & \ccell{53.6} & char\\
Pythia-6.7B & \ccell{36.3} & \ccell{39.5} & \ccell{39.0} & \ccell{44.6} & pmi & \ccell{71.4} & \ccell{72.6} & \ccell{70.0} & \ccell{64.1} & char & \ccell{68.7} & \ccell{66.9} & \ccell{68.7} & \ccell{47.6} & none\\
RPJ-INCITE-7B & \ccell{40.3} & \ccell{43.5} & \ccell{42.9} & \ccell{45.3} & pmi & \ccell{76.1} & \ccell{78.8} & \ccell{75.9} & \ccell{70.1} & char & \ccell{72.0} & \ccell{74.5} & \ccell{72.0} & \ccell{72.4} & char\\
MPT-7b & \ccell{41.7} & \ccell{46.3} & \ccell{44.7} & \ccell{45.7} & char & \ccell{76.3} & \ccell{78.0} & \ccell{76.2} & \ccell{68.5} & char & \ccell{82.4} & \ccell{79.9} & \ccell{82.4} & \ccell{76.7} & none\\
Falcon-7B & \ccell{41.6} & \ccell{47.4} & \ccell{47.6} & \ccell{49.7} & pmi & \ccell{77.0} & \ccell{80.6} & \ccell{78.3} & \ccell{69.8} & char & \ccell{78.2} & \ccell{78.8} & \ccell{78.2} & \ccell{77.6} & char\\
OLMo-7B & \ccell{41.6} & \ccell{45.5} & \ccell{45.0} & \ccell{46.4} & pmi & \ccell{76.7} & \ccell{78.9} & \ccell{77.4} & \ccell{69.6} & char & \ccell{78.7} & \ccell{77.7} & \ccell{78.7} & \ccell{78.6} & none\\
StableLM2-1.6B & \ccell{42.2} & \ccell{44.3} & \ccell{44.9} & \ccell{47.3} & pmi & \ccell{73.3} & \ccell{75.3} & \ccell{74.4} & \ccell{70.0} & char & \ccell{82.3} & \ccell{82.0} & \ccell{82.3} & \ccell{76.1} & none\\
Llama2-7B & \ccell{48.4} & \ccell{52.0} & \ccell{50.2} & \ccell{54.2} & pmi & \ccell{81.4} & \ccell{84.0} & \ccell{81.0} & \ccell{74.7} & char & \ccell{86.1} & \ccell{85.6} & \ccell{86.1} & \ccell{80.5} & none\\
OLMo-7B-0424 & \ccell{45.5} & \ccell{49.3} & \ccell{48.5} & \ccell{51.2} & pmi & \ccell{79.2} & \ccell{81.5} & \ccell{79.7} & \ccell{71.1} & char & \ccell{85.9} & \ccell{83.8} & \ccell{85.9} & \ccell{85.6} & none\\
Llama2-13B & \ccell{52.4} & \ccell{57.1} & \ccell{54.2} & \ccell{56.2} & char & \ccell{83.9} & \ccell{85.9} & \ccell{82.8} & \ccell{77.6} & char & \ccell{86.7} & \ccell{88.2} & \ccell{86.7} & \ccell{77.5} & char\\
Llama3-8B & \ccell{53.6} & \ccell{58.4} & \ccell{56.8} & \ccell{57.1} & char & \ccell{85.8} & \ccell{86.6} & \ccell{85.8} & \ccell{76.6} & char & \ccell{87.5} & \ccell{87.8} & \ccell{87.5} & \ccell{67.0} & char\\
Mistral-7B-v0.1 & \ccell{56.1} & \ccell{60.2} & \ccell{58.9} & \ccell{59.6} & char & \ccell{84.7} & \ccell{86.8} & \ccell{84.6} & \ccell{78.6} & char & \ccell{89.3} & \ccell{89.1} & \ccell{89.3} & \ccell{89.2} & none\\
Llama3-70B & \ccell{65.7} & \ccell{69.0} & \ccell{67.7} & \ccell{69.0} & char & \ccell{89.7} & \ccell{89.6} & \ccell{90.4} & \ccell{82.6} & tok & \ccell{91.2} & \ccell{90.4} & \ccell{91.2} & \ccell{91.7} & pmi\\
\midrule
{\bf average scores} & \ccell{43.7} & \ccell{47.3} & \ccell{46.4} & \ccell{49.0} & NA & \ccell{77.0} & \ccell{78.7} & \ccell{76.5} & \ccell{70.0} & NA & \ccell{77.4} & \ccell{77.8} & \ccell{77.4} & \ccell{70.5} & NA\\
\midrule
{\bf win percentage} & \ccell{0.0} & \ccell{33.3} & \ccell{0.0} & \ccell{66.7} & {\bf pmi} & \ccell{6.7} & \ccell{86.7} & \ccell{6.7} & \ccell{0.0} & {\bf char} & \ccell{46.7} & \ccell{46.7} & \ccell{0.0} & \ccell{6.7} & {\bf none}\\
\bottomrule
\end{tabular}
\setlength\tabcolsep{6pt} 
  \caption{Comparing \rc{} normalization schemes (part 1).}
  \label{tab:appendix-rc-norm-1}
\end{table*}

\begin{table*}
  \centering
  \small

\setlength\tabcolsep{2.5pt} 
\begin{tabular}{lccccc|ccccc|ccccc}
\toprule
 & \multicolumn{5}{c|}{\bf{CSQA}} & \multicolumn{5}{c|}{\bf{HSwag}} & \multicolumn{5}{c}{\bf{MMLU}}\\
model & none & char & tok & pmi & best & none & char & tok & pmi & best & none & char & tok & pmi & best\\
\midrule
Pythia-1B & \ccell{47.7} & \ccell{50.9} & \ccell{47.3} & \ccell{50.9} & char & \ccell{39.2} & \ccell{48.0} & \ccell{47.8} & \ccell{41.0} & char & \ccell{29.5} & \ccell{31.1} & \ccell{30.8} & \ccell{32.3} & pmi\\
OLMo-1B & \ccell{56.8} & \ccell{60.0} & \ccell{57.6} & \ccell{62.2} & pmi & \ccell{50.9} & \ccell{65.2} & \ccell{64.1} & \ccell{49.8} & char & \ccell{31.7} & \ccell{33.4} & \ccell{33.3} & \ccell{34.2} & pmi\\
TinyLlama-1.1B & \ccell{58.9} & \ccell{60.5} & \ccell{55.9} & \ccell{61.1} & pmi & \ccell{46.9} & \ccell{60.8} & \ccell{59.7} & \ccell{48.5} & char & \ccell{31.2} & \ccell{33.6} & \ccell{33.0} & \ccell{34.5} & pmi\\
Pythia-6.7B & \ccell{59.5} & \ccell{62.2} & \ccell{58.9} & \ccell{62.1} & char & \ccell{50.4} & \ccell{66.1} & \ccell{65.9} & \ccell{53.5} & char & \ccell{34.9} & \ccell{37.7} & \ccell{37.0} & \ccell{37.9} & pmi\\
RPJ-INCITE-7B & \ccell{67.7} & \ccell{69.4} & \ccell{67.2} & \ccell{69.2} & char & \ccell{55.7} & \ccell{72.8} & \ccell{71.8} & \ccell{60.6} & char & \ccell{37.4} & \ccell{40.1} & \ccell{40.0} & \ccell{40.9} & pmi\\
MPT-7b & \ccell{69.6} & \ccell{70.3} & \ccell{69.1} & \ccell{70.9} & pmi & \ccell{60.5} & \ccell{79.6} & \ccell{76.5} & \ccell{61.5} & char & \ccell{37.8} & \ccell{40.6} & \ccell{40.1} & \ccell{40.4} & char\\
Falcon-7B & \ccell{70.0} & \ccell{70.3} & \ccell{69.5} & \ccell{73.4} & pmi & \ccell{60.7} & \ccell{79.0} & \ccell{78.4} & \ccell{60.0} & char & \ccell{39.3} & \ccell{42.1} & \ccell{41.9} & \ccell{42.1} & char\\
OLMo-7B & \ccell{69.0} & \ccell{70.0} & \ccell{67.9} & \ccell{70.8} & pmi & \ccell{59.3} & \ccell{78.1} & \ccell{76.3} & \ccell{64.2} & char & \ccell{37.9} & \ccell{40.5} & \ccell{40.5} & \ccell{40.6} & pmi\\
StableLM2-1.6B & \ccell{63.6} & \ccell{66.3} & \ccell{65.6} & \ccell{68.2} & pmi & \ccell{54.7} & \ccell{70.3} & \ccell{69.7} & \ccell{56.4} & char & \ccell{35.2} & \ccell{37.1} & \ccell{37.1} & \ccell{38.6} & pmi\\
Llama2-7B & \ccell{70.5} & \ccell{72.7} & \ccell{68.4} & \ccell{74.2} & pmi & \ccell{61.9} & \ccell{78.9} & \ccell{77.1} & \ccell{64.4} & char & \ccell{42.0} & \ccell{44.4} & \ccell{43.9} & \ccell{44.8} & pmi\\
OLMo-7B-0424 & \ccell{71.6} & \ccell{63.5} & \ccell{59.0} & \ccell{70.4} & none & \ccell{61.4} & \ccell{80.1} & \ccell{77.7} & \ccell{65.2} & char & \ccell{39.9} & \ccell{42.4} & \ccell{42.2} & \ccell{41.8} & char\\
Llama2-13B & \ccell{72.2} & \ccell{72.7} & \ccell{68.4} & \ccell{74.0} & pmi & \ccell{63.7} & \ccell{83.9} & \ccell{81.0} & \ccell{70.3} & char & \ccell{44.3} & \ccell{47.6} & \ccell{46.7} & \ccell{47.1} & char\\
Llama3-8B & \ccell{72.0} & \ccell{74.2} & \ccell{73.5} & \ccell{69.9} & char & \ccell{62.8} & \ccell{81.8} & \ccell{80.3} & \ccell{71.1} & char & \ccell{47.5} & \ccell{51.1} & \ccell{50.8} & \ccell{49.6} & char\\
Mistral-7B-v0.1 & \ccell{73.1} & \ccell{73.8} & \ccell{74.4} & \ccell{72.3} & tok & \ccell{64.5} & \ccell{83.0} & \ccell{81.0} & \ccell{70.3} & char & \ccell{46.9} & \ccell{50.3} & \ccell{50.0} & \ccell{49.0} & char\\
Llama3-70B & \ccell{77.1} & \ccell{77.1} & \ccell{77.1} & \ccell{75.8} & char & \ccell{70.3} & \ccell{89.5} & \ccell{87.1} & \ccell{80.8} & char & \ccell{57.2} & \ccell{60.7} & \ccell{60.5} & \ccell{59.4} & char\\
\midrule
{\bf average scores} & \ccell{66.6} & \ccell{67.6} & \ccell{65.3} & \ccell{68.4} & NA & \ccell{57.5} & \ccell{74.5} & \ccell{73.0} & \ccell{61.2} & NA & \ccell{39.5} & \ccell{42.2} & \ccell{41.9} & \ccell{42.2} & NA\\
\midrule
{\bf win percentage} & \ccell{6.7} & \ccell{33.3} & \ccell{6.7} & \ccell{53.3} & {\bf pmi} & \ccell{0.0} & \ccell{100.0} & \ccell{0.0} & \ccell{0.0} & {\bf char} & \ccell{0.0} & \ccell{46.7} & \ccell{0.0} & \ccell{53.3} & {\bf pmi}\\
\bottomrule
\end{tabular}
\setlength\tabcolsep{6pt} 
  \caption{Comparing \rc{} normalization schemes (part 2)}
  \label{tab:appendix-rc-norm-2}
\end{table*}

\begin{table*}
  \centering
  \small

\setlength\tabcolsep{2.5pt} 
\begin{tabular}{lccccc|ccccc|ccccc}
\toprule
 & \multicolumn{5}{c|}{\bf{OBQA}} & \multicolumn{5}{c|}{\bf{PIQA}} & \multicolumn{5}{c}{\bf{SIQA}}\\
model & none & char & tok & pmi & best & none & char & tok & pmi & best & none & char & tok & pmi & best\\
\midrule
Pythia-1B & \ccell{20.2} & \ccell{28.6} & \ccell{30.4} & \ccell{40.4} & pmi & \ccell{70.3} & \ccell{68.9} & \ccell{68.8} & \ccell{60.1} & none & \ccell{42.8} & \ccell{46.4} & \ccell{46.0} & \ccell{44.4} & char\\
OLMo-1B & \ccell{26.0} & \ccell{33.0} & \ccell{38.4} & \ccell{47.6} & pmi & \ccell{73.2} & \ccell{74.1} & \ccell{73.2} & \ccell{59.9} & char & \ccell{45.3} & \ccell{51.5} & \ccell{49.9} & \ccell{47.3} & char\\
TinyLlama-1.1B & \ccell{24.4} & \ccell{34.8} & \ccell{35.8} & \ccell{45.0} & pmi & \ccell{72.1} & \ccell{71.7} & \ccell{72.3} & \ccell{62.0} & tok & \ccell{45.6} & \ccell{50.4} & \ccell{48.2} & \ccell{48.4} & char\\
Pythia-6.7B & \ccell{25.8} & \ccell{37.0} & \ccell{37.4} & \ccell{50.4} & pmi & \ccell{74.8} & \ccell{74.9} & \ccell{74.3} & \ccell{63.6} & char & \ccell{48.0} & \ccell{51.7} & \ccell{52.8} & \ccell{49.2} & tok\\
RPJ-INCITE-7B & \ccell{31.8} & \ccell{40.0} & \ccell{42.8} & \ccell{49.0} & pmi & \ccell{74.9} & \ccell{75.9} & \ccell{76.0} & \ccell{61.9} & tok & \ccell{50.8} & \ccell{56.6} & \ccell{56.0} & \ccell{52.2} & char\\
MPT-7b & \ccell{31.6} & \ccell{43.8} & \ccell{43.8} & \ccell{52.4} & pmi & \ccell{77.7} & \ccell{79.2} & \ccell{78.1} & \ccell{63.7} & char & \ccell{51.0} & \ccell{57.4} & \ccell{55.9} & \ccell{52.5} & char\\
Falcon-7B & \ccell{35.2} & \ccell{45.8} & \ccell{44.4} & \ccell{55.2} & pmi & \ccell{78.3} & \ccell{79.0} & \ccell{79.2} & \ccell{63.2} & tok & \ccell{52.9} & \ccell{60.1} & \ccell{57.5} & \ccell{54.4} & char\\
OLMo-7B & \ccell{33.2} & \ccell{42.8} & \ccell{45.0} & \ccell{55.8} & pmi & \ccell{78.2} & \ccell{78.5} & \ccell{79.3} & \ccell{65.2} & tok & \ccell{50.3} & \ccell{56.5} & \ccell{56.5} & \ccell{52.8} & char\\
StableLM2-1.6B & \ccell{34.4} & \ccell{41.6} & \ccell{45.2} & \ccell{51.0} & pmi & \ccell{75.2} & \ccell{75.6} & \ccell{75.9} & \ccell{63.6} & tok & \ccell{52.7} & \ccell{61.1} & \ccell{60.7} & \ccell{56.1} & char\\
Llama2-7B & \ccell{33.8} & \ccell{44.6} & \ccell{45.0} & \ccell{57.8} & pmi & \ccell{76.7} & \ccell{77.5} & \ccell{77.7} & \ccell{62.9} & tok & \ccell{52.6} & \ccell{59.6} & \ccell{58.3} & \ccell{53.6} & char\\
OLMo-7B-0424 & \ccell{37.2} & \ccell{48.4} & \ccell{49.6} & \ccell{59.8} & pmi & \ccell{78.5} & \ccell{80.3} & \ccell{79.3} & \ccell{66.3} & char & \ccell{53.5} & \ccell{54.9} & \ccell{54.3} & \ccell{55.7} & pmi\\
Llama2-13B & \ccell{39.2} & \ccell{46.4} & \ccell{48.4} & \ccell{60.8} & pmi & \ccell{78.9} & \ccell{80.2} & \ccell{79.8} & \ccell{66.4} & char & \ccell{56.7} & \ccell{63.6} & \ccell{60.7} & \ccell{56.8} & char\\
Llama3-8B & \ccell{37.0} & \ccell{47.6} & \ccell{50.0} & \ccell{56.2} & pmi & \ccell{79.7} & \ccell{81.6} & \ccell{81.1} & \ccell{67.5} & char & \ccell{54.6} & \ccell{62.6} & \ccell{60.1} & \ccell{56.4} & char\\
Mistral-7B-v0.1 & \ccell{38.2} & \ccell{48.4} & \ccell{50.0} & \ccell{61.0} & pmi & \ccell{80.8} & \ccell{82.8} & \ccell{81.3} & \ccell{67.4} & char & \ccell{55.6} & \ccell{63.0} & \ccell{60.9} & \ccell{57.5} & char\\
Llama3-70B & \ccell{47.0} & \ccell{55.0} & \ccell{56.6} & \ccell{69.0} & pmi & \ccell{82.8} & \ccell{83.1} & \ccell{83.2} & \ccell{68.3} & tok & \ccell{59.7} & \ccell{65.6} & \ccell{64.8} & \ccell{57.3} & char\\
\midrule
{\bf average scores} & \ccell{33.0} & \ccell{42.5} & \ccell{44.2} & \ccell{54.1} & NA & \ccell{76.8} & \ccell{77.6} & \ccell{77.3} & \ccell{64.1} & NA & \ccell{51.5} & \ccell{57.4} & \ccell{56.2} & \ccell{53.0} & NA\\
\midrule
{\bf win percentage} & \ccell{0.0} & \ccell{0.0} & \ccell{0.0} & \ccell{100.0} & {\bf pmi} & \ccell{6.7} & \ccell{46.7} & \ccell{46.7} & \ccell{0.0} & {\bf char} & \ccell{0.0} & \ccell{86.7} & \ccell{6.7} & \ccell{6.7} & {\bf char}\\
\bottomrule
\end{tabular}
\setlength\tabcolsep{6pt} 
  \caption{Comparing \rc{} normalization schemes (part 3).}
  \label{tab:appendix-rc-norm-3}
\end{table*}

\subsubsection{Tasks that generally prefer \rc{}}
\label{rc-tasks-appendix}
\task{HellaSwag} and \task{WinoGrande} continue to have \rc{} scores higher than \mc{} scores even for the strongest models that can understand the \mc{} prompt. This somewhat surprising tendency seems correlated with the fact that these tasks in the \rc{} format are exactly like the language modeling task of finding the most natural continuation of a running piece of text. Judging from the trends in the plot, it would also be interesting to monitor if as even more capable models are developed, the \mc{} scores will eventually surpass that of the \rc{} scores (given how close they already get to each other).

\subsubsection{Hybrid formulation}
\label{subsubsec:hybrid_formulation}
In \rc{}, overall probability score could be quite misleading since it may heavily favor shorter answers with fewer tokens. Note that this would be different if the answer choices are actually listed before scoring the answer string, then most tokens (after the choice has been disambiguated by the first few tokens) would have probability near one. This ``hybrid'' formulation has been used in some cases, but usually scores in between the \rc{} and \mc{} approaches \citep{wiegreffe-etal-2023-increasing}. However, this hybrid approach is not popular in evaluation standardization efforts like the Open LLM Leaderboard, HELM, or when used to evaluate models during development, so it is not a focus in \evalstandard{}.

\subsection{Tokenization of MCQA choice labels}
\label{tokenization-details-appendix}

When formatting multiple-choice questions, \evalstandard{} specifies the use of a prefix space in front of each answer choice, that is \prompt{"\textbackslash n~A.~<choice>"} rather than \prompt{"\textbackslash nA.~<choice>"}. Figure~\ref{fig:appendix-tokenizer-spacing} shows explicit examples of tokenizers where this helps maintain a correspondence between the token for the answer label and the token in the final answer (e.g., \prompt{"\textbackslash nAnswer: A"}). E.g., for the Llama tokenizer, the consistent token is the \prompt{"\_A"} rather than the separate token \prompt{"A"} you get without the prefix space.

\begin{figure*}[h]
\fcolorbox{black}{lightgray}{
\begin{minipage}{0.95\linewidth}{\tt \frenchspacing \small
> from transformers import AutoTokenizer\\
> llama\_tokenizer = AutoTokenizer.from\_pretrained("meta-llama/Llama-2-7b-hf")\\
> olmo\_tokenizer = AutoTokenizer.from\_pretrained("allenai/OLMo-7B-0424-hf")\\
> test\_string = "What is 3+4?\textbackslash n A. 7\textbackslash nA. 7\textbackslash nAnswer: A"\\
> llama\_tokenizer.tokenizer(test\_string)\\
{[}'\_What', '\_is', '\_', '3', '+', '4', '?', '<0x0A>', '\_A', '.', '\_', '7', '<0x0A>', 'A', '.', '\_', '7', '<0x0A>', 'Answer', ':', '\_A'] \\
> olmo\_tokenizer.tokenizer(test\_string)\\
{[}'What', 'Ġis', 'Ġ3', '+', '4', '?', 'Ċ', 'ĠA', '.', 'Ġ7', 'Ċ', 'A', '.', 'Ġ7', 'Ċ', 'Answer', ':', 'ĠA']
}
 \end{minipage}
}
\caption{Tokenizer example, showing two examples of tokenizers which need a prefix space before MCQA answer choice labels to represent the choice label and the final answer label using the same token.}
\label{fig:appendix-tokenizer-spacing}
\end{figure*}

\section{Extended \evalstandard{} result table}
\label{appendix-extended-results}

Table~\ref{score-table-main-extended} shows \evalstandard{} evaluations across an extended set of 40 models. Table~\ref{score-table-variations-obqa} shows an extended version of Table~\ref{score-table-variations-arc-c} which includes score variations across different references on \task{OpenbookQA} in addition to \task{ARC-Challenge}.

\begin{table*}
\setlength\tabcolsep{2pt} 
  \centering
\begin{normalsize}
\begin{tabular}{lccccccccccc}
\toprule
\small{\bf{model}} & \small{\bf{ARC\_C}} & \small{\bf{ARC\_E}} & \small{\bf{BoolQ}} & \small{\bf{CSQA}} & \small{\bf{HSwag}} & \small{\bf{MMLU}} & \small{\bf{OBQA}} & \small{\bf{PIQA}} & \small{\bf{SIQA}} & \small{\bf{WinoG}} & \small{\bf{average}}\\
\midrule
Pythia-1B & \ccell{31.4} & \ccell{63.4} & $\ccell{56.8}^\dagger$ & \ccell{50.9} & \ccell{48.0} & \ccell{31.1} & \ccell{40.4} & \ccell{68.9} & \ccell{46.4} & \ccell{52.7} & \ccell{49.0}\\
OLMo-1B-0724 & \ccell{36.4} & \ccell{53.5} & \ccell{66.8} & \ccell{42.4} & \ccell{67.5} & \ccell{32.0} & \ccell{44.2} & \ccell{74.0} & \ccell{45.2} & \ccell{62.9} & \ccell{52.5}\\
OLMo-1B & \ccell{38.6} & \ccell{68.3} & \ccell{51.3} & \ccell{62.2} & \ccell{65.2} & \ccell{33.4} & \ccell{47.6} & \ccell{74.1} & \ccell{51.5} & \ccell{59.3} & \ccell{55.1}\\
TinyLlama-1.1B & \ccell{38.1} & \ccell{69.5} & \ccell{63.6} & \ccell{61.1} & \ccell{60.8} & \ccell{33.6} & \ccell{45.0} & \ccell{71.7} & \ccell{50.4} & \ccell{60.1} & \ccell{55.4}\\
Qwen2-0.5B & $\ccell{48.4}^\dagger$ & $\ccell{64.9}^\dagger$ & \ccell{64.3} & \ccell{56.2} & \ccell{48.9} & $\ccell{45.3}^\dagger$ & $\ccell{51.6}^\dagger$ & \ccell{67.9} & $\ccell{54.7}^\dagger$ & \ccell{56.1} & \ccell{55.8}\\
Llama3.2-1B & \ccell{43.5} & \ccell{71.6} & \ccell{69.4} & \ccell{59.6} & \ccell{67.3} & \ccell{38.2} & \ccell{42.0} & \ccell{73.7} & \ccell{52.0} & \ccell{62.5} & \ccell{58.0}\\
Pythia-6.7B & \ccell{44.6} & \ccell{72.6} & \ccell{68.7} & \ccell{62.1} & \ccell{66.1} & \ccell{37.7} & \ccell{50.4} & \ccell{74.9} & \ccell{51.7} & \ccell{62.3} & \ccell{59.1}\\
RPJ-INCITE-7B & \ccell{45.3} & \ccell{78.8} & \ccell{72.0} & \ccell{69.2} & \ccell{72.8} & \ccell{40.1} & \ccell{49.0} & \ccell{75.9} & \ccell{56.6} & \ccell{68.0} & \ccell{62.8}\\
Gemma-2B & \ccell{49.9} & \ccell{80.2} & \ccell{76.6} & \ccell{68.9} & \ccell{72.5} & $\ccell{41.7}^\dagger$ & \ccell{52.4} & \ccell{76.1} & \ccell{57.1} & \ccell{66.1} & \ccell{64.2}\\
StableLM2-1.6B & $\ccell{50.6}^\dagger$ & \ccell{75.3} & \ccell{82.3} & $\ccell{70.4}^\dagger$ & \ccell{70.3} & $\ccell{40.4}^\dagger$ & $\ccell{56.6}^\dagger$ & \ccell{75.6} & $\ccell{64.3}^\dagger$ & \ccell{65.7} & \ccell{65.1}\\
OLMo-7B & \ccell{46.4} & \ccell{78.9} & \ccell{78.7} & \ccell{70.8} & \ccell{78.1} & \ccell{40.5} & \ccell{55.8} & \ccell{78.5} & \ccell{56.5} & \ccell{68.5} & \ccell{65.3}\\
MPT-7b & \ccell{45.7} & \ccell{78.0} & \ccell{82.4} & \ccell{70.9} & \ccell{79.6} & \ccell{40.6} & \ccell{52.4} & \ccell{79.2} & \ccell{57.4} & \ccell{70.2} & \ccell{65.6}\\
Zamba2-1.2B & $\ccell{55.0}^\dagger$ & \ccell{85.4} & \ccell{76.1} & \ccell{70.1} & \ccell{73.4} & $\ccell{44.7}^\dagger$ & $\ccell{59.8}^\dagger$ & \ccell{76.6} & \ccell{58.4} & \ccell{67.2} & \ccell{66.7}\\
Falcon-7B & \ccell{49.7} & \ccell{80.6} & \ccell{78.2} & \ccell{73.4} & \ccell{79.0} & \ccell{42.1} & \ccell{55.2} & \ccell{79.0} & \ccell{60.1} & \ccell{71.3} & \ccell{66.9}\\
DCLM-1B & $\ccell{57.6}^\dagger$ & \ccell{79.5} & \ccell{80.9} & \ccell{71.3} & \ccell{75.1} & $\ccell{48.5}^\dagger$ & $\ccell{60.0}^\dagger$ & \ccell{76.6} & $\ccell{60.5}^\dagger$ & \ccell{68.1} & \ccell{67.8}\\
DeepSeek-MoE-16B & \ccell{53.4} & \ccell{82.7} & \ccell{81.9} & \ccell{72.7} & \ccell{80.4} & $\ccell{45.5}^\dagger$ & \ccell{58.4} & \ccell{80.1} & \ccell{59.9} & \ccell{73.2} & \ccell{68.8}\\
Llama2-7B & \ccell{54.2} & \ccell{84.0} & \ccell{86.1} & \ccell{74.2} & \ccell{78.9} & $\ccell{46.2}^\dagger$ & \ccell{57.8} & \ccell{77.5} & \ccell{59.6} & \ccell{71.7} & \ccell{69.0}\\
DeepSeek-7B & $\ccell{57.2}^\dagger$ & \ccell{80.6} & \ccell{84.8} & \ccell{74.0} & \ccell{80.4} & $\ccell{48.7}^\dagger$ & $\ccell{62.2}^\dagger$ & \ccell{79.3} & $\ccell{65.1}^\dagger$ & \ccell{72.5} & \ccell{70.5}\\
Qwen2-1.5B & $\ccell{68.6}^\dagger$ & $\ccell{85.2}^\dagger$ & \ccell{75.3} & $\ccell{72.0}^\dagger$ & \ccell{67.6} & $\ccell{56.5}^\dagger$ & $\ccell{74.6}^\dagger$ & \ccell{75.7} & $\ccell{65.3}^\dagger$ & \ccell{64.5} & \ccell{70.5}\\
OLMoE-1B-7B-0924 & $\ccell{62.1}^\dagger$ & \ccell{84.2} & \ccell{79.2} & \ccell{72.9} & \ccell{80.0} & $\ccell{54.1}^\dagger$ & $\ccell{65.4}^\dagger$ & \ccell{79.8} & $\ccell{63.0}^\dagger$ & \ccell{70.2} & \ccell{71.1}\\
Gemma2-2B & $\ccell{67.5}^\dagger$ & $\ccell{84.3}^\dagger$ & \ccell{83.6} & $\ccell{66.4}^\dagger$ & \ccell{74.6} & $\ccell{53.3}^\dagger$ & $\ccell{68.8}^\dagger$ & \ccell{78.5} & $\ccell{64.7}^\dagger$ & \ccell{71.8} & \ccell{71.3}\\
Llama3.2-3B & $\ccell{69.6}^\dagger$ & $\ccell{85.1}^\dagger$ & \ccell{78.3} & \ccell{69.0} & \ccell{77.0} & $\ccell{57.8}^\dagger$ & $\ccell{67.2}^\dagger$ & \ccell{77.4} & $\ccell{64.9}^\dagger$ & \ccell{69.9} & \ccell{71.6}\\
JetMoE-8B & $\ccell{61.4}^\dagger$ & $\ccell{81.9}^\dagger$ & \ccell{85.7} & $\ccell{75.3}^\dagger$ & \ccell{81.7} & $\ccell{49.1}^\dagger$ & $\ccell{68.0}^\dagger$ & \ccell{80.3} & $\ccell{71.3}^\dagger$ & \ccell{70.7} & \ccell{72.5}\\
Llama2-13B & $\ccell{67.3}^\dagger$ & \ccell{85.9} & \ccell{86.7} & \ccell{74.0} & \ccell{83.9} & $\ccell{55.8}^\dagger$ & $\ccell{65.4}^\dagger$ & \ccell{80.2} & $\ccell{65.9}^\dagger$ & \ccell{74.9} & \ccell{74.0}\\
OLMo-7B-0424 & $\ccell{66.9}^\dagger$ & $\ccell{83.6}^\dagger$ & \ccell{85.9} & $\ccell{85.8}^\dagger$ & \ccell{80.1} & $\ccell{54.4}^\dagger$ & $\ccell{68.6}^\dagger$ & \ccell{80.3} & $\ccell{76.1}^\dagger$ & \ccell{73.6} & \ccell{75.5}\\
OLMo-7B-0724 & $\ccell{68.0}^\dagger$ & $\ccell{85.7}^\dagger$ & \ccell{85.3} & $\ccell{85.4}^\dagger$ & \ccell{80.5} & $\ccell{54.9}^\dagger$ & $\ccell{67.6}^\dagger$ & \ccell{79.3} & $\ccell{76.1}^\dagger$ & \ccell{73.2} & \ccell{75.6}\\
DeepSeek-V2-Lite & $\ccell{74.0}^\dagger$ & $\ccell{88.9}^\dagger$ & \ccell{84.7} & \ccell{73.8} & \ccell{81.9} & $\ccell{58.8}^\dagger$ & $\ccell{72.4}^\dagger$ & \ccell{80.2} & $\ccell{69.1}^\dagger$ & \ccell{74.0} & \ccell{75.8}\\
Qwen1.5-MoE-A2.7B & $\ccell{77.4}^\dagger$ & $\ccell{91.6}^\dagger$ & \ccell{85.0} & $\ccell{81.4}^\dagger$ & \ccell{80.0} & $\ccell{62.4}^\dagger$ & $\ccell{80.6}^\dagger$ & \ccell{81.0} & $\ccell{74.1}^\dagger$ & \ccell{72.3} & \ccell{78.6}\\
Llama3-8B & $\ccell{79.3}^\dagger$ & $\ccell{92.4}^\dagger$ & \ccell{87.5} & $\ccell{73.9}^\dagger$ & \ccell{81.8} & $\ccell{66.6}^\dagger$ & $\ccell{77.2}^\dagger$ & \ccell{81.6} & $\ccell{70.2}^\dagger$ & \ccell{76.2} & \ccell{78.7}\\
Mistral-7B-v0.3 & $\ccell{78.3}^\dagger$ & $\ccell{91.1}^\dagger$ & \ccell{88.4} & $\ccell{72.7}^\dagger$ & \ccell{83.1} & $\ccell{63.5}^\dagger$ & $\ccell{80.0}^\dagger$ & \ccell{81.9} & $\ccell{71.2}^\dagger$ & \ccell{77.7} & \ccell{78.8}\\
Llama3.1-8B & $\ccell{79.5}^\dagger$ & $\ccell{91.7}^\dagger$ & \ccell{88.5} & $\ccell{74.3}^\dagger$ & \ccell{81.6} & $\ccell{66.9}^\dagger$ & $\ccell{78.6}^\dagger$ & \ccell{81.1} & $\ccell{71.4}^\dagger$ & \ccell{76.6} & \ccell{79.0}\\
Mistral-7B-v0.1 & $\ccell{78.6}^\dagger$ & $\ccell{90.8}^\dagger$ & \ccell{89.3} & $\ccell{72.4}^\dagger$ & \ccell{83.0} & $\ccell{64.0}^\dagger$ & $\ccell{80.6}^\dagger$ & \ccell{82.8} & $\ccell{71.3}^\dagger$ & \ccell{77.9} & \ccell{79.1}\\
DCLM-7B & $\ccell{79.8}^\dagger$ & $\ccell{92.3}^\dagger$ & \ccell{87.0} & \ccell{77.0} & \ccell{82.3} & $\ccell{64.4}^\dagger$ & $\ccell{79.6}^\dagger$ & \ccell{80.1} & $\ccell{71.2}^\dagger$ & \ccell{77.3} & \ccell{79.1}\\
Qwen2-7B & $\ccell{88.1}^\dagger$ & $\ccell{95.3}^\dagger$ & \ccell{88.9} & $\ccell{81.2}^\dagger$ & $\ccell{86.4}^\dagger$ & $\ccell{71.8}^\dagger$ & $\ccell{88.2}^\dagger$ & $\ccell{86.0}^\dagger$ & $\ccell{78.0}^\dagger$ & \ccell{75.1} & \ccell{83.9}\\
Gemma2-9B & $\ccell{89.5}^\dagger$ & $\ccell{95.5}^\dagger$ & \ccell{89.4} & $\ccell{78.8}^\dagger$ & $\ccell{87.3}^\dagger$ & $\ccell{70.6}^\dagger$ & $\ccell{88.4}^\dagger$ & $\ccell{86.1}^\dagger$ & $\ccell{76.0}^\dagger$ & \ccell{78.8} & \ccell{84.0}\\
Mixtral-8x7B-v0.1 & $\ccell{87.1}^\dagger$ & $\ccell{96.1}^\dagger$ & $\ccell{90.0}^\dagger$ & $\ccell{78.3}^\dagger$ & \ccell{86.7} & $\ccell{71.9}^\dagger$ & $\ccell{87.0}^\dagger$ & $\ccell{86.1}^\dagger$ & $\ccell{75.1}^\dagger$ & \ccell{82.6} & \ccell{84.1}\\
Zamba2-7B & $\ccell{92.2}^\dagger$ & $\ccell{96.7}^\dagger$ & \ccell{89.3} & $\ccell{84.0}^\dagger$ & $\ccell{89.4}^\dagger$ & $\ccell{68.5}^\dagger$ & $\ccell{84.2}^\dagger$ & $\ccell{86.5}^\dagger$ & $\ccell{77.7}^\dagger$ & \ccell{79.6} & \ccell{84.8}\\
Llama3.1-70B & $\ccell{92.8}^\dagger$ & $\ccell{97.4}^\dagger$ & \ccell{91.9} & $\ccell{81.7}^\dagger$ & \ccell{89.4} & $\ccell{79.1}^\dagger$ & $\ccell{92.6}^\dagger$ & $\ccell{91.2}^\dagger$ & $\ccell{80.6}^\dagger$ & \ccell{84.5} & \ccell{88.1}\\
Llama3-70B & $\ccell{93.7}^\dagger$ & $\ccell{97.7}^\dagger$ & $\ccell{91.7}^\dagger$ & $\ccell{83.2}^\dagger$ & \ccell{89.5} & $\ccell{79.8}^\dagger$ & $\ccell{93.4}^\dagger$ & $\ccell{91.6}^\dagger$ & $\ccell{78.9}^\dagger$ & \ccell{84.1} & \ccell{88.4}\\
Qwen2.5-72B & $\ccell{95.5}^\dagger$ & $\ccell{98.8}^\dagger$ & $\ccell{91.9}^\dagger$ & $\ccell{89.7}^\dagger$ & $\ccell{97.5}^\dagger$ & $\ccell{85.3}^\dagger$ & $\ccell{97.4}^\dagger$ & $\ccell{94.0}^\dagger$ & $\ccell{82.2}^\dagger$ & $\ccell{84.3}^\dagger$ & \ccell{91.7}\\
\bottomrule
\end{tabular}
\end{normalsize}
\setlength\tabcolsep{6pt} 
  \caption{Extended reproducible performance scores across models and tasks using \evalstandard{}, providing robust, meaningful comparisons across a wide range of models and tasks. $^\dagger$ indicates use of the \mc\ score. 
  }
  \label{score-table-main-extended}
\end{table*}

\begin{table}[t]
\centering
\setlength\tabcolsep{1.5pt} 
\begin{small}

\begin{tabular}{l|ccccccc|c|cccccc}
\toprule
 & \multicolumn{7}{c|}{{\bf \task{ARC-Challenge} Evaluations:}} & \ \ \  & \multicolumn{6}{c}{\bf \task{OpenbookQA} Evaluations:} \\
{\bf Model$\downarrow$} & {\bf Ref1} & {\bf Ref2} & {\bf Ref3} & {\bf Ref4} & {\bf Ref5} & {\bf Ref6} & {\bf OLMES} & {\bf } & {\bf Ref2} & {\bf Ref4} & {\bf Ref5} & {\bf Ref7} & {\bf Ref8} & {\bf OLMES}\\
\midrule
MPT-7B & \ccell{47.7} & \ccell{42.6} &  &  & \ccell{46.5} &  & \ccell{45.7} &  & \ccell{51.4} &  & \ccell{48.6} &  &  & \ccell{52.4}\\
RPJ-INCITE-7B & \ccell{46.3} &  &  &  & \ccell{42.8} &  & \ccell{45.3} &  &  &  & \ccell{49.4} &  &  & \ccell{49.0}\\
Falcon-7B & \ccell{47.9} & \ccell{42.4} &  & \ccell{44.5} & \ccell{47.5} &  & \ccell{49.7} &  & \ccell{51.6} & \ccell{44.6} & \ccell{53.0} &  & $\ccell{26.0}^\dagger$ & \ccell{55.2}\\
Mistral-7B & \ccell{60.0} &  & \ccell{55.5} & \ccell{54.9} &  &  & $\ccell{78.6}^\dagger$ &  &  &  &  & \ccell{52.2} & $\ccell{77.6}^\dagger$ & $\ccell{80.6}^\dagger$\\
Llama2-7B & \ccell{53.1} & \ccell{45.9} & \ccell{43.2} & \ccell{45.9} & \ccell{48.5} & $\ccell{53.7}^\dagger$ & \ccell{54.2} &  & \ccell{58.6} & \ccell{58.6} & \ccell{48.4} & \ccell{58.6} & $\ccell{54.4}^\dagger$ & \ccell{57.8}\\
Llama2-13B & \ccell{59.4} & \ccell{49.4} & \ccell{48.8} & \ccell{49.4} &  & $\ccell{67.6}^\dagger$ & $\ccell{67.3}^\dagger$ &  & \ccell{57.0} & \ccell{57.0} &  & \ccell{57.0} & $\ccell{63.4}^\dagger$ & $\ccell{65.4}^\dagger$\\
Llama3-8B & \ccell{60.2} &  &  &  &  & $\ccell{78.6}^\dagger$ & $\ccell{79.3}^\dagger$ &  &  &  &  &  & $\ccell{76.6}^\dagger$ & $\ccell{77.2}^\dagger$\\
\midrule
Num shots & 25 & 0 & 0 & 0 & 0 & 25 & 5 &  & 0 & 0 & 0 & 0 & 5 & 5\\
Curated shots & No &  &  &  &  & No & Yes &  &  &  &  &  & No & Yes\\
Formulation & \rc{} & \rc{} & \rc{}? & \rc{} & \rc{} & \mc{} & \mc{}/\rc{} &  & \rc{} & \rc{} & \rc{} & \rc{} & \mc{} & \mc{}/\rc{}\\
Normalization & char & char & ? & char? & pmi & none & none/pmi &  & pmi & pmi? & pmi & pmi? & none & none/pmi\\
\bottomrule
\end{tabular}
\end{small}
\\
\vspace{2mm}
\begin{small}
\begin{tabular}{lllll}
\toprule
\textbf{Ref} & \textbf{Reference citation} & \ \ \ \ & \textbf{Ref} & \textbf{Reference citation} \\
\midrule
\textbf{Ref1} & HF Open LLM Leaderboard \citep{open-llm-leaderboard} & & \textbf{Ref5} & OLMo paper \citep{groeneveld2024olmo}\\
\textbf{Ref2} & Llama2 paper \citep{touvron2023llama} & & \textbf{Ref6} & Llama3 model card \citep{llama3modelcard}\\
\textbf{Ref3} & Mistral 7B \citep{jiang2023mistral} & &  \textbf{Ref7} & Gemma paper \citep{gemmateam2024gemma}\\
\textbf{Ref4} & Falcon paper \citep{almazrouei2023falcon} & & \textbf{Ref8} & HELM Lite Leaderboard \citep{liang2023holistic} \\
\bottomrule
\end{tabular} 
\end{small}
\caption{Extended version of Table~\ref{score-table-variations-arc-c} showing scores reported in different references for LLM performances on \task{ARC-Challenge} and \task{OpenbookQA}. Scores indicated with $^\dagger$ are using multiple-choice formulation (\mc{}) rather than  ``cloze'' formulation (\rc{}) (see Section~\ref{sec:mcqa-llm} for definitions). Entries with ``?'' denote either undocumented or mixed approaches across models. Different references use different evaluation setups, some of which are not fully specified, so conclusions about which models perform best are not reproducible.}
  \label{score-table-variations-obqa}

\setlength\tabcolsep{6pt} 
\end{table}

\section{HELM Reproduction of MMLU}
\label{appendix-helm-mmlu-repro}
\begin{figure}[t!]
\centering
      \includegraphics[width=\columnwidth]{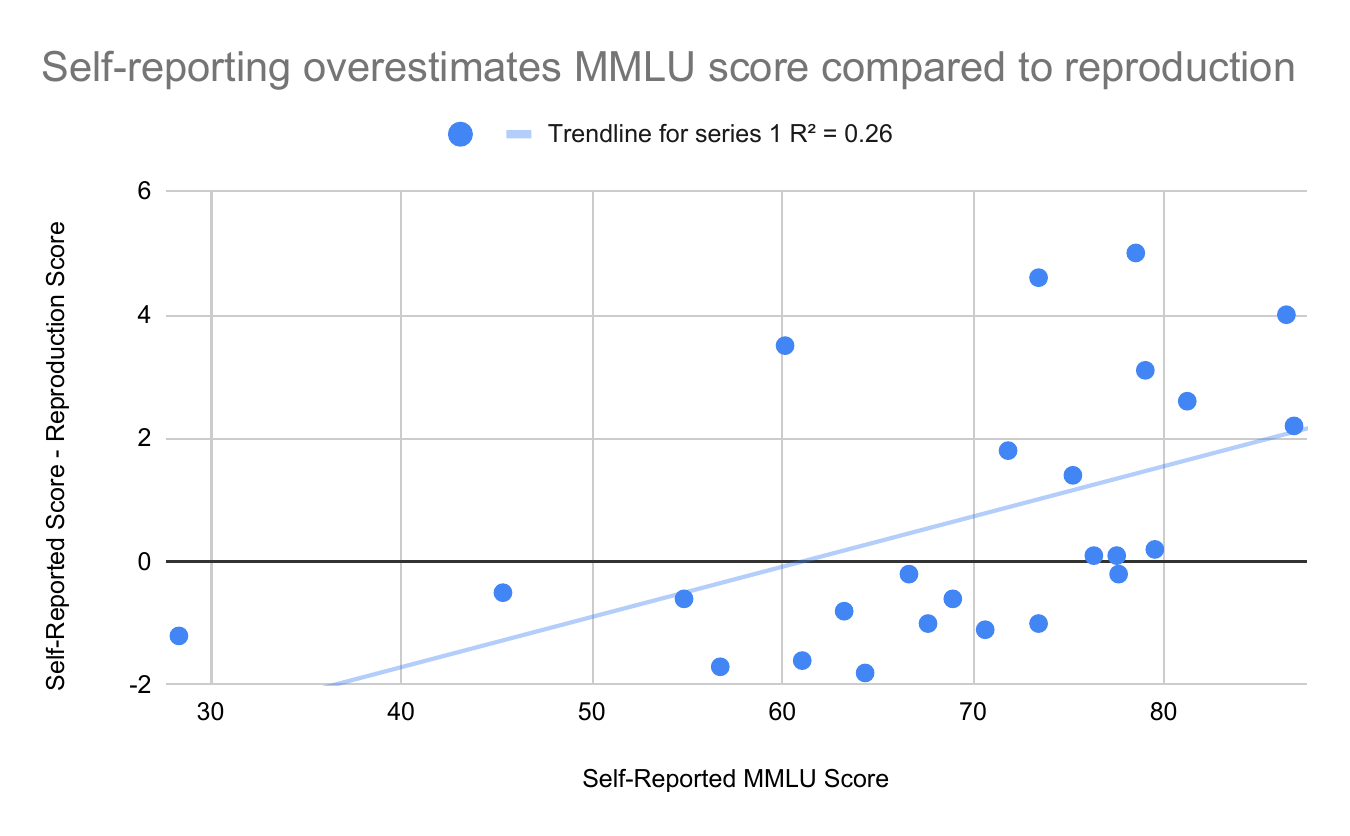}
\caption{Self-reporting overestimates \task{MMLU} score compared to reproduction, from \url{https://crfm.stanford.edu/2024/05/01/helm-mmlu.html}. Each point corresponds to a model, the x-axis shows self-reported \task{MMLU} score, and the y-axis shows the difference between the self-reported score and the reproduced score. Points above the y=0 line have higher self-reported performance than the reproduction; the trend line has a positive slope, indicating that on average, the higher the self-reported score the more they overestimate performance compared to the reproduction.}
\label{fig:helm_mmlu_reproduction}
\end{figure}

In Figure~\ref{fig:helm_mmlu_reproduction} we see data taken from HELM's reproduction of MMLU scores for a variety of models. 

\section{Compute used}
\label{compute-used}
The inference on the models evaluated were done on 	NVIDIA RTX A6000 GPUs. A total of around 400 GPU hours was used.


\section{Curation of 5-shot examples: considerations}
\label{curation_5_shot_considerations}
Procedure for manually curating the few-shot examples:

\begin{itemize}
    \item Download the train set from Hugging Face datasets 
    \item Start from the beginning of the training set, looking at a batch of 10 (i.e., start with first 10)
    \item Skip ambiguous instances
    \item Skip instances that hint at discrimination or otherwise deemed inappropriate
    \item Skip instances if the same label has appeared frequently (e.g., 4 consecutive instances with gold label `C', keep better ones out of those)
    \item If instances are grouped/labeled by topic, choose instances to be diverse (e.g., first 3 are all about a certain topic, pick from later ones to ensure diversity).
    \item If you end up with less than 7 instances that cover the label space or range of different topics, look at the next batch of 10.
    \item Finally, reorder instances to obtain a somewhat balanced output of answer labels -- the first 5 shots should cover the space of answer labels.
\end{itemize}

Note that a few more than 5 shots per dataset were curated in the process, though in practice we are just using the first 5.

\section{\evalstandard{} prompt formats for each task}
\label{task-prompt-formats-appendix}
In Figure~\ref{prompt:ARC-Challenge-RC-5shot} we show an example of a full 5-shot prompt from \task{ARC-Challenge} (\rc{}). Then we show single instance formatting for each of the 10 tasks in Figures~\ref{prompt:ARC-Challenge-MC}-
\ref{prompt:Winogrande}. For each task, we show both the \mc{} and \rc{} formats.

All curated few-shot examples and prompt formatting code are available by accessing \evalstandardurl{}.

\begin{figure} 
{\frenchspacing \tt \footnotesize \begin{tabularx}{\linewidth}{r X} \toprule Prompt & Question: George wants to warm his hands quickly by rubbing them. Which skin surface will produce the most heat?\\
 & Answer: dry palms\\
 & \\
 & Question: Which of the following statements best explains why magnets usually stick to a refrigerator door?\\
 & Answer: The refrigerator door contains iron.\\
 & \\
 & Question: A fold observed in layers of sedimentary rock most likely resulted from the\\
 & Answer: converging of crustal plates.\\
 & \\
 & Question: Which of these do scientists offer as the most recent explanation as to why many plants and animals died out at the end of the Mesozoic era?\\
 & Answer: impact of an asteroid created dust that blocked the sunlight\\
 & \\
 & Question: Which of the following is a trait that a dog does NOT inherit from its parents?\\
 & Answer: the size of its appetite\\
 & \\
 & Question: A boat is acted on by a river current flowing north and by wind blowing on its sails. The boat travels northeast. In which direction is the wind most likely applying force to the sails of the boat?\\
 & Answer: \\ 
 \midrule Completion & \ east \\ 
 \bottomrule 
\end{tabularx} } 
\caption{\evalstandard{} 5-shot prompt example for \task{ARC-Challenge} (\rc{}).}  \label{prompt:ARC-Challenge-RC-5shot}
\end{figure}

\begin{figure} 
{\frenchspacing \tt \footnotesize \begin{tabularx}{\linewidth}{r X} \toprule Prompt & Question: George wants to warm his hands quickly by rubbing them. Which skin surface will produce the most heat?\\
 & \ A. dry palms\\
 & \ B. wet palms\\
 & \ C. palms covered with oil\\
 & \ D. palms covered with lotion\\
 & Answer: \\ 
 \midrule Completion & \ A \\ 
 \bottomrule 
\end{tabularx} } 
\caption{\evalstandard{} prompt example for \task{ARC-Challenge} (\mc{}).}  \label{prompt:ARC-Challenge-MC}
\end{figure}

\begin{figure} 
{\frenchspacing \tt \footnotesize \begin{tabularx}{\linewidth}{r X} \toprule Prompt & Question: George wants to warm his hands quickly by rubbing them. Which skin surface will produce the most heat?\\
 & Answer: \\ 
 \midrule Completion & \ dry palms \\ 
 \bottomrule 
\end{tabularx} } 
\caption{\evalstandard{} prompt example for \task{ARC-Challenge} (\rc{}).}  \label{prompt:ARC-Challenge}
\end{figure}

\begin{figure} 
{\frenchspacing \tt \footnotesize \begin{tabularx}{\linewidth}{r X} \toprule Prompt & Question: Lichens are symbiotic organisms made of green algae and fungi. What do the green algae supply to the fungi in this symbiotic relationship?\\
 & \ A. carbon dioxide\\
 & \ B. food\\
 & \ C. protection\\
 & \ D. water\\
 & Answer: \\ 
 \midrule Completion & \ B \\ 
 \bottomrule 
\end{tabularx} } 
\caption{\evalstandard{} prompt example for \task{ARC-Easy} (\mc{}).}  \label{prompt:ARC-Easy-MC}
\end{figure}

\begin{figure} 
{\frenchspacing \tt \footnotesize \begin{tabularx}{\linewidth}{r X} \toprule Prompt & Question: Lichens are symbiotic organisms made of green algae and fungi. What do the green algae supply to the fungi in this symbiotic relationship?\\
 & Answer: \\ 
 \midrule Completion & \ food \\ 
 \bottomrule 
\end{tabularx} } 
\caption{\evalstandard{} prompt example for \task{ARC-Easy} (\rc{}).}  \label{prompt:ARC-Easy}
\end{figure}

\begin{figure} 
{\frenchspacing \tt \footnotesize \begin{tabularx}{\linewidth}{r X} \toprule Prompt & Persian language -- Persian, also known by its endonym Farsi, is one of the Western Iranian languages within the Indo-Iranian branch of the Indo-European language family. It is primarily spoken in Iran, Afghanistan (officially known as Dari since 1958), and Tajikistan (officially known as Tajiki since the Soviet era), and some other regions which historically were Persianate societies and considered part of Greater Iran. It is written in the Persian alphabet, a modified variant of the Arabic script, which itself evolved from the Aramaic alphabet.\\
 & Question: do iran and afghanistan speak the same language?\\
 & \ A. yes\\
 & \ B. no\\
 & Answer: \\ 
 \midrule Completion & \ A \\ 
 \bottomrule 
\end{tabularx} } 
\caption{\evalstandard{} prompt example for \task{BoolQ} (\mc{}).}  \label{prompt:BoolQ-MC}
\end{figure}

\begin{figure} 
{\frenchspacing \tt \footnotesize \begin{tabularx}{\linewidth}{r X} \toprule Prompt & Persian language -- Persian, also known by its endonym Farsi, is one of the Western Iranian languages within the Indo-Iranian branch of the Indo-European language family. It is primarily spoken in Iran, Afghanistan (officially known as Dari since 1958), and Tajikistan (officially known as Tajiki since the Soviet era), and some other regions which historically were Persianate societies and considered part of Greater Iran. It is written in the Persian alphabet, a modified variant of the Arabic script, which itself evolved from the Aramaic alphabet.\\
 & Question: do iran and afghanistan speak the same language?\\
 & Answer: \\ 
 \midrule Completion & \ yes \\ 
 \bottomrule 
\end{tabularx} } 
\caption{\evalstandard{} prompt example for \task{BoolQ} (\rc{}).}  \label{prompt:BoolQ}
\end{figure}

\begin{figure} 
{\frenchspacing \tt \footnotesize \begin{tabularx}{\linewidth}{r X} \toprule Prompt & Question: Sammy wanted to go to where the people were.  Where might he go?\\
 & \ A. race track\\
 & \ B. populated areas\\
 & \ C. the desert\\
 & \ D. apartment\\
 & \ E. roadblock\\
 & Answer: \\ 
 \midrule Completion & \ B \\ 
 \bottomrule 
\end{tabularx} } 
\caption{\evalstandard{} prompt example for \task{CommonsenseQA} (\mc{}).}  \label{prompt:CommonsenseQA-MC}
\end{figure}

\begin{figure} 
{\frenchspacing \tt \footnotesize \begin{tabularx}{\linewidth}{r X} \toprule Prompt & Question: Sammy wanted to go to where the people were.  Where might he go?\\
 & Answer: \\ 
 \midrule Completion & \ populated areas \\ 
 \bottomrule 
\end{tabularx} } 
\caption{\evalstandard{} prompt example for \task{CommonsenseQA} (\rc{}).}  \label{prompt:CommonsenseQA}
\end{figure}

\begin{figure} 
{\frenchspacing \tt \footnotesize \begin{tabularx}{\linewidth}{r X} \toprule Prompt & Health: How to cope with suicidal thoughts. Put off any plans. Promise yourself that you'll wait 48 hours before doing anything. Remember, thoughts don't have the power to force you to act. \\
 & Choose the best continuation:\\
 & \ A. Even when you do, there may be a small image of the future still lurking around your brain. For instance, don't tell yourself that you can't make it.\\
 & \ B. You're doing something, and no one can force you to act. It's completely natural to feel negative thoughts before you act.\\
 & \ C. Do not panic if people talk to you (even if it's about quitting smoking). Have a plan for how you're going to react to a group of people who bring on suicidal thoughts.\\
 & \ D. Sometimes extreme pain can distort our perception. Waiting before taking action will give your mind time to clear.\\
 & Answer: \\ 
 \midrule Completion & \ D \\ 
 \bottomrule 
\end{tabularx} } 
\caption{\evalstandard{} prompt example for \task{HellaSwag} (\mc{}).}  \label{prompt:HellaSwag-MC}
\end{figure}

\begin{figure} 
{\frenchspacing \tt \footnotesize \begin{tabularx}{\linewidth}{r X} \toprule Prompt & Health: How to cope with suicidal thoughts. Put off any plans. Promise yourself that you'll wait 48 hours before doing anything. Remember, thoughts don't have the power to force you to act.\\ 
 \midrule Completion & \ Sometimes extreme pain can distort our perception. Waiting before taking action will give your mind time to clear.  \\ 
 \bottomrule 
\end{tabularx} } 
\caption{\evalstandard{} prompt example for \task{HellaSwag} (\rc{}).}  \label{prompt:HellaSwag}
\end{figure}

\begin{figure} 
{\frenchspacing \tt \footnotesize \begin{tabularx}{\linewidth}{r X} \toprule 
Instruction & The following are multiple choice questions (with answers) about abstract algebra.\\
Prompt & Question: Find all c in Z\_3 such that Z\_3[x]/(x\textasciicircum2 + c) is a field.\\
 & \ A. 0\\
 & \ B. 1\\
 & \ C. 2\\
 & \ D. 3\\
 & Answer: \\ 
 \midrule Completion & \ B \\ 
 \bottomrule 
\end{tabularx} } 
\caption{\evalstandard{} prompt example for \task{MMLU} (abstract\_algebra) (\mc{}).}  \label{prompt:MMLU (abstract_algebra)-MC}
\end{figure}

\begin{figure} 
{\frenchspacing \tt \footnotesize \begin{tabularx}{\linewidth}{r X} \toprule 
Instruction & The following are multiple choice questions (with answers) about abstract algebra.\\
Prompt & Question: Find all c in Z\_3 such that Z\_3[x]/(x\textasciicircum2 + c) is a field.\\
 & Answer: \\ 
 \midrule Completion & \ 1 \\ 
 \bottomrule 
\end{tabularx} } 
\caption{\evalstandard{} prompt example for \task{MMLU} (abstract\_algebra) (\rc{}).}  \label{prompt:MMLU (abstract_algebra)}
\end{figure}

\begin{figure} 
{\frenchspacing \tt \footnotesize \begin{tabularx}{\linewidth}{r X} \toprule Prompt & Question: When standing miles away from Mount Rushmore\\
 & \ A. the mountains seem very close\\
 & \ B. the mountains are boring\\
 & \ C. the mountains look the same as from up close\\
 & \ D. the mountains seem smaller than in photographs\\
 & Answer: \\ 
 \midrule Completion & \ D \\ 
 \bottomrule 
\end{tabularx} } 
\caption{\evalstandard{} prompt example for \task{OpenbookQA} (\mc{}).}  \label{prompt:OpenbookQA-MC}
\end{figure}

\begin{figure} 
{\frenchspacing \tt \footnotesize \begin{tabularx}{\linewidth}{r X} \toprule Prompt & Question: When standing miles away from Mount Rushmore\\
 & Answer: \\ 
 \midrule Completion & \ the mountains seem smaller than in photographs \\ 
 \bottomrule 
\end{tabularx} } 
\caption{\evalstandard{} prompt example for \task{OpenbookQA} (\rc{}).}  \label{prompt:OpenbookQA}
\end{figure}

\begin{figure} 
{\frenchspacing \tt \footnotesize \begin{tabularx}{\linewidth}{r X} \toprule Prompt & Goal: how do you stab something?\\
 & \ A. stick a sharp object through it.\\
 & \ B. pin it with a sharp object.\\
 & Answer: \\ 
 \midrule Completion & \ A \\ 
 \bottomrule 
\end{tabularx} } 
\caption{\evalstandard{} prompt example for Physical Interaction QA (\mc{}).}  \label{prompt:Physical Interaction QA-MC}
\end{figure}

\begin{figure} 
{\frenchspacing \tt \footnotesize \begin{tabularx}{\linewidth}{r X} \toprule Prompt & Goal: how do you stab something?\\
 & Answer: \\ 
 \midrule Completion & \ stick a sharp object through it. \\ 
 \bottomrule 
\end{tabularx} } 
\caption{\evalstandard{} prompt example for Physical Interaction QA (\rc{}).}  \label{prompt:Physical Interaction QA}
\end{figure}

\begin{figure} 
{\frenchspacing \tt \footnotesize \begin{tabularx}{\linewidth}{r X} \toprule Prompt & Question: Cameron decided to have a barbecue and gathered her friends together. How would Others feel as a result?\\
 & \ A. like attending\\
 & \ B. like staying home\\
 & \ C. a good friend to have\\
 & Answer: \\ 
 \midrule Completion & \ A \\ 
 \bottomrule 
\end{tabularx} } 
\caption{\evalstandard{} prompt example for \task{Social IQa} (\mc{}).}  \label{prompt:Social IQa-MC}
\end{figure}

\begin{figure} 
{\frenchspacing \tt \footnotesize \begin{tabularx}{\linewidth}{r X} \toprule Prompt & Question: Cameron decided to have a barbecue and gathered her friends together. How would Others feel as a result?\\
 & Answer: \\ 
 \midrule Completion & \ like attending \\ 
 \bottomrule 
\end{tabularx} } 
\caption{\evalstandard{} prompt example for \task{Social IQa} (\rc{}).}  \label{prompt:Social IQa}
\end{figure}

\begin{figure} 
{\frenchspacing \tt \footnotesize \begin{tabularx}{\linewidth}{r X} \toprule Prompt & Fill in the blank: John moved the couch from the garage to the backyard to create space. The \_\_\_ is small.\\
 & \ A. garage\\
 & \ B. backyard\\
 & Answer: \\ 
 \midrule Completion & \ A \\ 
 \bottomrule 
\end{tabularx} } 
\caption{\evalstandard{} prompt example for \task{WinoGrande} (\mc{}).}  \label{prompt:Winogrande-MC}
\end{figure}

\begin{figure} 
{\frenchspacing \tt \footnotesize \begin{tabularx}{\linewidth}{r X} \toprule Prompt1 & John moved the couch from the garage to the backyard to create space. The garage\\ 
Prompt2 & John moved the couch from the garage to the backyard to create space. The backyard\\ 
 \midrule Completion & \  is small. \\ 
 \bottomrule 
\end{tabularx} } 
\caption{\evalstandard{} prompt example for \task{WinoGrande} (\rc{}). In this case the completions are the same for each answer choice, but the prompt is different.}  \label{prompt:Winogrande}
\end{figure}
\end{document}